\documentclass[twocolumn]{svjour3}
\usepackage{multirow}
\usepackage{pdfpages}
\usepackage{algorithm}
\usepackage[noend]{algorithmic}
\usepackage[round]{natbib}
\usepackage{color}
\smartqed
\RequirePackage{fix-cm}

\definecolor{purple}{RGB}{100,0,100}

\begin{document}

\title{Hybrid Ant Colony Optimization in solving Multi--Skill Resource--Constrained Project Scheduling Problem}

\author{Pawe\l \space B. Myszkowski \and Marek E. Skowro\'nski \and \L ukasz P. Olech \and Krzysztof O\'sliz\l o}
\institute{Department of Artificial Intelligence, Wroclaw University of Technology, \email{\{pawel.myszkowski, m.e.skowronski\}@pwr.wroc.pl, \{179214, 163753\}@student.pwr.wroc.pl}}
 
\maketitle

\begin{abstract}
In this paper Hybrid Ant Colony Optimization (HAntCO) approach in solving Multi--Skill Resource Constrained Project Scheduling Problem (MS--RCPSP) has been presented. We have proposed hybrid approach that links classical heuristic priority rules for project scheduling with Ant Colony Optimization (ACO). Furthermore, a novel approach for updating pheromone value has been proposed, based on both the best and worst solutions stored by ants. The objective of this paper is to research the usability and robustness of ACO and its hybrids with priority rules in solving MS--RCPSP. Experiments have been performed using artificially created dataset instances, based on real--world ones. We published those instances that can be used as a benchmark. Presented results show that ACO--based hybrid method is an efficient approach. More directed search process by hybrids makes this approach more stable and provides mostly better results than classical ACO.

\end{abstract}

\section{Introduction}

Resource--Constrained Project Scheduling Problem \\ (RCPSP) is one of the most investigated types of scheduling problems. Its goal is to find the resource--to--task assignments to make the finite project plan the cheapest or shortest. Description of RCPSP in \citep{blazewicz} as combinatorial, NP--hard problem encouraged scientists to find \textit{good enough} methods that would be able to produce approximate, (sub)optimal solutions in finite, polynomial computing time. Those methods are called (meta)heuristics and are used to solve problems for which finding optimal solution in an acceptable time is impossible. 

Beside Evolutionary Algorithms (EA), Taboo Search (TS), Simulated Annealing (SA) and some other techniques, metaheuristics contain also a group of methods called \textit{swarm--intelligence} methods, as Particle Swarm Optimization (PSO) or Ant Colony Optimization (ACO). Those methods assume that separate individuals, representing given problem solutions, can interact with each other and cooperate to achieve their common goals. In this point of view, swarm intelligence techniques are similar to EA. However, they assume that there is one, constant population of individuals that can evolve in time but cannot be replaced by new individuals.  ACO, as the name stands, simulates the behaviour of ants, travelling between the ant's nest and the source of food. The optimization goal is to find the optimal path between food and nest, while definition of \textit{path's quality} is varied and dependent on the considered problem. 

The \textit{real--life} nature of RCPSP comes from business. Project managers in companies struggle to build effective project schedule, meeting duration, cost and other constraints. What is more, many constraints have to be satisfied, while manual scheduling often leads to violating of those constraints. It is a common problem for project managers. Hence computer--aided, (semi--)automatic tools are desired by the industry. Furthermore, obtaining the project plan by computer--driven methods is less time--consuming than obtained manually. 

Developing RCPSP to a more practical problem, we have introduced the skills domain, transforming it to the Multi--Skill RCPSP (MS--RCPSP) extension. In MS--RCPSP resources dispose of some given pool of skills, while every task requires some skill in a given level to be performed. It means not every resource is capable of performing every task. As solution space in MS--RCPSP is more constrained, it is more difficult to build \textit{good enough} solution -- project schedule. Furthermore, we have added another criterion -- project schedule performance cost, transforming the classical single--objective (duration) RCPSP into multi--objective (duration vs. cost) MS--RCPSP.

We have decided to create hybrid methods by combining ACO--based approach with some heuristics described in \citep{skowronski_heu}. Therefore, classical heuristics have been also investigated. Based on results obtained in that paper, we have chosen given heuristics that could be used to obtain the initial solution for ACO mechanism and stand as a Hybrid Ant Colony Optimization (HAntCO). A very significant fact is that depending on optimized criterion (duration or cost) various priority rule could be used. Therefore, we are able to decide whether using HAntCO allows to get better solutions than using ACO mechanism not supported by any priority rule. 

Investigating ACO--based approach was motivated by the willingness to compare results obtained using several collective intelligence methods and other metaheuristics, such as TS or SA \citep{skowronski_ts} to solve this problem. As we had researched EA--based approach before \citep{skowronski_men}, we made a comparison of different approaches in case of their robustness, effectiveness and stability, while those terms would be explained further. 

The dataset for experiments has been created artificially, but instances are based on the real--world ones obtained from an international enterprise. What is more, presented MS--RCPSP could be generalized to the PSP-LIB \citep{kolisch} dataset model that is regarded as a benchmark for methods solving project scheduling problems. 

The rest of the paper is organized as follows. Section \ref{sec:related_work} describes selected ways of solving the (MS--) RCPSP using metaheuristics, especially ACO. Section \ref{sec:statement} presents the MS--RCPSP problem statement, while Section \ref{sec:approach} describes the approaches proposed in this paper. Section \ref{sec:experiments} provides conducted experiments of proposed methods in a given dataset. Finally, section \ref{sec:conclusions} presents the conclusions of obtained results and suggests some directions of future work.


\section{Related work \label{sec:related_work}}

Metaheuristics are very often used to solve RCPSP because of its NP--hard nature. EA (\citep{hartmann2}, \citep{valls}, \citep{valls2}), TS (\citep{thomas}, \citep{tsai}, \citep{verhoeven}), SA (\citep{bouleimen}, \citep{das}) are well explored and widely applied to solve MS--RCPSP. It is worth a mention that ACO is not the only swarm intelligence metaheuristic used in solving (MS--) RCPSP. PSO approaches could be found in \citep{tam}, \citep{zhang3}, \citep{zhang4}, while bee colony optimization (BCO) method has been investigated in \citep{ziarati}. Numerous papers regarding PSO or BCO in solving RCPSP prove that those methods are often investigated and researched. 

However, there is still lack of papers regarding multi--objective Multi--Skill extension of RCPSP. Some approaches solving MS--RCPSP in project duration domain \citep{alanzi}, \citep{santos} or project cost domain \citep{li} could be found. On the other hand, there are methods solving classical RCPSP extended by cost domain but without skills considerations. Such research has been presented in \\ \citep{phru}, \citep{jaberi}, \citep{gonzalez}, \citep{luna} and \citep{yannibelli}. Hence we have decided to combine those two elements: multi--objective optimization and multi--skill domain for project scheduling problem. 

Although classical RCPSP is deeply investigated and numerous approaches could be easily compared using PSPLIB instances, it is very hard to find multi--objective MS--RCPSP methods working on datasets that could be regarded as a benchmark. Some papers describe instances artificially generated (\citep{hegazy}, \citep{santos}), while some others propose methods of PSPLIB dataset adaptation (\citep{alanzi}, \citep{drezet}, \citep{kadrou}, \citep{li}). However, both of those approaches for handling MS--RCPSP benchmark data are not supplied by any published dataset instances. Hence the need of proposing our own dataset has arisen. 

ACO is inspired by the rules in the real environment of ants. Real ants are capable of finding the shortest path from the source of food to the ant's nest. Every ant from a population leaves a substance called pheromone while getting to the source of food. This substance attracts other ants to come into that direction. However, the pheromone evaporates gradually in every period. It means the shorter path is, the less pheromone would be evaporated and that path would be more attractive to other ants. In that way, more and more ants start to exploit the region of a surface where there was more pheromone -- the path to the source of food was shorter. Finally, all ants move along the same path, what is regarded as the found solution of the problem. 

A classical ACO approach with some modifications that made it more robust has been presented in \citep{merkle}. Particularly, the following features have been proposed: combination of two pheromone updating methods, dynamic influence of those methods during ACO runtime and possibility of leaving the best obtained solution by an elitist ant to preserve sticking in local optima. The presented methods have been tested on PSPLIB instances. In many cases, the obtained results were better than the best found so far, what confirms the robustness of that approach. 

Various improvements of ACO have been proposed in \citep{luo}. A single solution, represented by a single ant, is obtained using serial generation scheme. If generated schedule turns out to be infeasible after adding a given task, the ant can reschedule some beginning fragment of a current schedule in order to make it feasible. The feasibility is lost when precedence constraints are violated. The following activities that should be added to a current schedule are chosen by combination of classical heuristics: most total successors, latest finish time (LFT)  and resource scheduling method. The authors used UBO dataset from ProGen \citep{kolisch} to verify their approach.

A different ACO approach has been presented in \citep{zhou2} as well. The combination of Ant Colony System \citep{dorigo} and Max--Min Ant System \citep{stutzle} called MMACS has been proposed. The following improvements have been proposed in this approach: pseudorandom proportional rule for choosing a next activity, updating pheromone only in the base of the best ant from given iteration and serial schedule generation scheme. Furthermore, an extended and RCPSP--adjusted 2opt local search method \citep{watson} called PS--2opt has been proposed. Results of experiments conducted on PSPLIB stated that PS--2opt and MMACS methods are robust in solving RCPSP.

Another ACO--based approach has been presented in \citep{liang} where activity--on--node task precedence relations representation is considered. Activity selection is performed by forward--parallel method, while the search space exploration and exploitation is performed by tuned online and offline pheromone updating procedure. Conclusions supported by performed experiments on PSPLIB datasets stand that the approach proposed in \citep{liang} gives competitive results in comparison to other (not--only) ACO--based approaches.


\section{Problem statement \label{sec:statement}}

Before the description of the multi--skill extension for RCPSP, the fundamentals of classical RCPSP would be presented. The motivation to investigate RCPSP and its extensions came from industry and would be explained in detail in Subsection \ref{subsec:adjustment}.

\subsection{\textit{Classical RCPSP description}}
In RCPSP a set of tasks is given, while every task is described by its duration, start and finish dates. Tasks are non-preemptive. It means any task cannot be withdrawn if it has been started. Tasks are related to each other by precedence relations, describing which tasks are needed to be completed before some other could be started. Tasks that have to be finished before the start time of another task are called predecessors. In classical RCPSP resource units are provided. Every resource owns a finite number of units (represented as integer numbers) that could be assigned to various tasks while tasks require some number of units to be performed. Cumulative number of units of tasks assigned to specified resource in a given period cannot exceed a number of units owned by resource. Only one resource can be assigned to a given task but not only one task can be assigned to given resource in given timestamp. In classical RCPSP, two dummy activities are added - start and finish tasks. It is because, in RCPSP, every task besides the start one has predecessors. Hence finish time of the last, dummy finish task is the finish time of schedule and the duration of a project could be computed as duration between start time of dummy start task and finish time of finish dummy task. The goal of RCPSP is to find such task--to--resource assignments to make the final schedule feasible and as shortest as possible. Combinatorial nature of the RCPSP makes it NP--hard. 

A solution of RCPSP is a feasible schedule -- the one in which resource units and precedence constraints are preserved. 

\subsection{\textit{Multi--skill extension of RCPSP}}
\begin{figure}[!ht]
\centering
\includegraphics[scale=0.65]{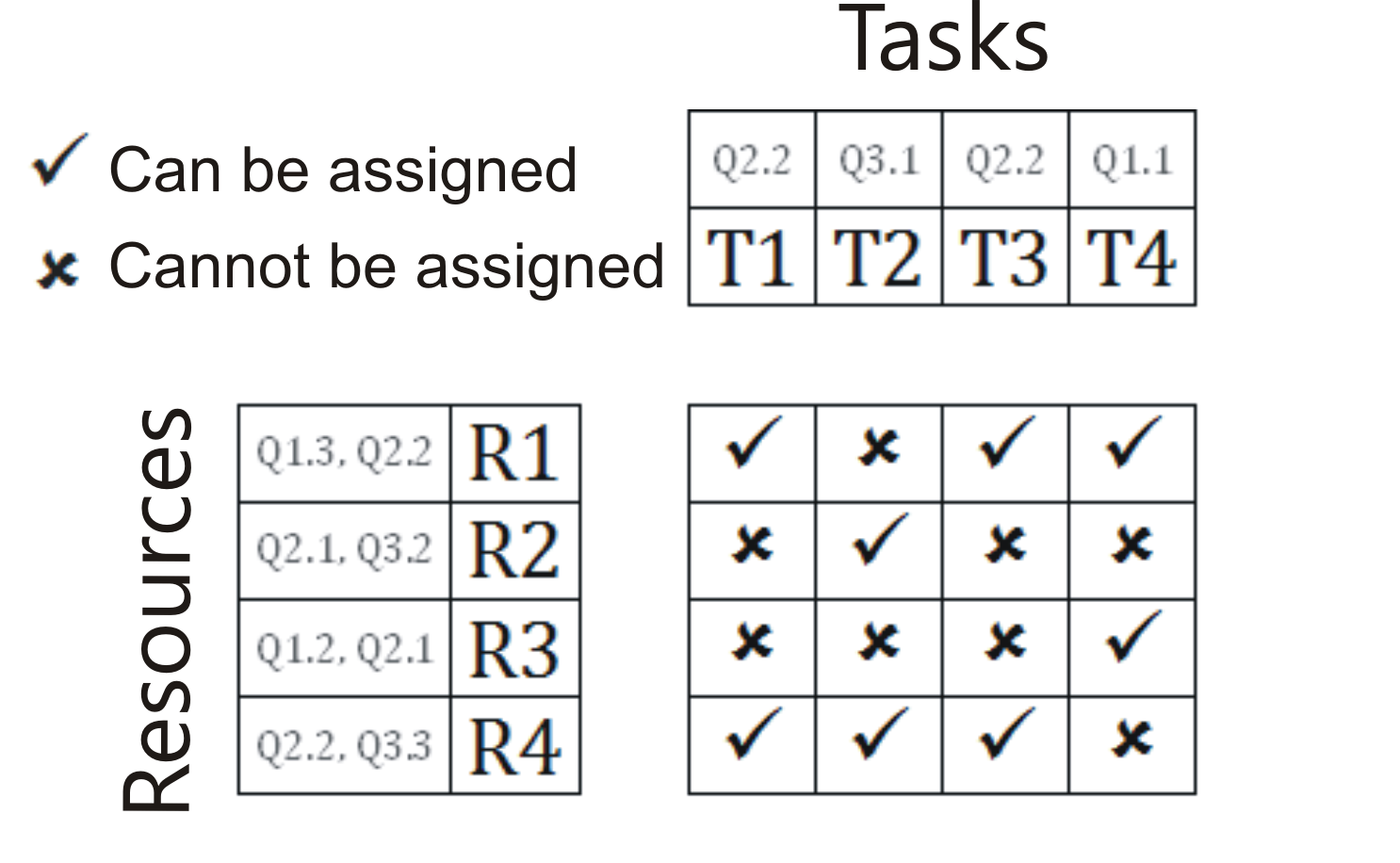}
\caption{Example of skill matrix}
\label{img:matrix}
\end{figure}

MS--RCPSP extension adds the skills domain to classical RCPSP. Every task requires some skill at given familiarity level to be performed, while every resource disposes some skills pool -- subset of skill types (e.g. developer, analyst, tester, architect, etc.) defined in a project with given familiarity level. Therefore, the resource $R$ is capable of performing the task $T$ only if $R$ disposes skill required by $T$ at the same or higher level. The capabilities of performing tasks by resources could be presented as skill matrix. Sample skill matrix has been shown in the Fig. \ref{img:matrix}.

In the skill matrix presented in the Fig. \ref{img:matrix} skills required by task to be performed have been written over task definition, while skills owned by resources have been written next to resource definition. This figure presents sample resource capabilities: resource $R1$ disposes skills $Q1$ and $Q2$ with familiarity level 3 and 2 respectively. It is capable of performing task $T1$, $T3$ and $T4$ because all of those mentioned tasks require skill owned by $R1$ at no higher level than it has. $R1$ cannot be assigned to $T2$, because this task requires totally different skill that $R1$ does not dispose of, even at the lowest familiarity level. Analogously, resource $R2$ can be assigned to task $T2$, resource $R3$ is a proper one for task $T3$ and, finally, resource $R4$ can perform tasks $T1$, $T2$ and $T3$. Even though $R3$ disposes of skill $Q2$, it cannot be assigned to $T1$ and $T3$ because those tasks require $Q2$ at higher familiarity level that this resource disposes. 

\subsection{\textit{Model adjustment}}\label{subsec:adjustment}
As a result of consultations with representatives of various enterprises, we decided to introduce some practical changes in classical RCPSP extended to MS--RCPSP model. Firstly, we introduced resource salary (as an hourly wage) paid for performed work. In that case, resources are regarded only as human ones varied by their salary. We also resigned from introducing start and finish dummy activities as our approach assumes that there could be some tasks that are not connected by precedence relations with any other. Hence we cannot define the project duration, start time and finish time based on dummy activities. 

What is more, resources are not described by units -- any resource cannot be assigned to more than one task in an overlapping period -- dedicated resources \citep{bianco}. If such a situation occurs, the \textbf{conflict} is detected and should be resolved. The conflict fixing procedure is presented in Subsec. \ref{subsec:conflict}. 

%

Schedule feasibility for such modified problem is extended from classical RCPSP schedule feasibility definition by skills domain -- only resources capable of performing given tasks can be assigned to them. 

\subsection{\textit{Problem formulation}}
Feasible Project Schedule ($PS$) consists of $J=1,...,n$ tasks and $K=1,...,m$ resources. A non pre--emptive duration $d_j$, start time $S_j$ and finish time $F_j$ is defined for each task. Predecessors of given task $j$ are defined as $P_j$. Each resource is defined by its hourly rate salary $s_k$ and owned skills $Q^k=1,...,r$, while pool of owned skills is a subset of all skills defined in project $Q^k \in Q$. Value $l_q$ denotes the level of given skill, while $h_q$ describes its type and $q_j$ is a skill required by $j$ to be performed. Therefore, by $J^k$ subset of tasks that can be performed by $k$ resource is defined. Duration of a project schedule is denoted as $\tau$. Cost of performing $j$ task by $k$ resource is denoted as $c_j^k=d_j*s_k$, where $s_k$ describes the salary of resource $k$ assigned to $j$. For simplicity, we have modified the task's performance cost from $c_j^k$ to $c_j$, because only one resource can be assigned to given task. Hence there is no need to distinguish various costs for the same task. Moreover, we have introduced variable defining whether $k$ is assigned to $j$ in given time $t$: $U_{j,k}^t \in \{0;1\}$. If $U_{j,k}^t=1$, $k$ is assigned to $j$ in $t$. Analogously, $k$ is not assigned to $j$ in $t$ if $U_{j,k}^t = 0$. 

Feasible project schedule ($PS$) belongs to the set of all feasible and non--feasible solutions (violating pre-cedence-, resource- and skills-constraints) : $PS \in PS_{all}$.

Formally, the problem could be regarded as optimization (minimization) problem and stated as follows:
\begin{equation}\label{eq:dur_opt}
\min{f(PS)} = \min{[f_{\tau}(PS),f_{C}(PS)]}
\end{equation}
Subject to:
\begin{equation}\label{eq:cost}
\forall_{k \in K} s_k \geq 0 , \forall_{k \in K} Q^k \neq \emptyset
\end{equation}
\begin{equation}\label{eq:finish_times}
\forall_{j \in J} F_j \geq 0 ; \forall_{j \in J} d_j \geq 0
\end{equation}
\begin{equation}\label{eq:prec_constr}
\forall_{j \in J, j \neq 1, i \in P_j} F_i \leq F_j-d_j
\end{equation}
\begin{equation}\label{eq:skill_constr}
\forall_{i \in J^k} \; \exists_{q \in Q^k} \; h_q = h_{q_i} \wedge l_q \geq l_{q_i}
\end{equation}
\begin{equation}\label{eq:once_in_time}
\forall_{k \in K} \forall_{t \in \tau} \sum_{i=1}^n U_{i,k}^t \leq 1
\end{equation}
\begin{equation}\label{eq:each_task}
\forall_{j \in J} \exists_{!t \in \tau, !k \in K}   U_{j,k}^t = 1
\end{equation}
Eq. \ref{eq:dur_opt} denotes the duration and cost optimization respectively. Depending on the evaluation function configuration (described below), various optimization modes could be used in an optimization process. $f_{\tau}(PS)$ is an evaluation function of project schedule's duration, while $f_{C}(PS)$ is an evaluation function of project schedule's performance cost. 

The first constraint (Eq. \ref{eq:cost}) preserves the positive values of resource salaries and ability to perform at least one task by every resource. Eq. \ref{eq:finish_times} states that every task has positive finish date and duration, while Eq. \ref{eq:prec_constr} shows the precedence constrains rule. Next two equations: Eq. \ref{eq:skill_constr} introduces skill constraints and transforms RCPSP into MS--RCPSP. Constraint (Eq. \ref{eq:once_in_time}) describes that any resource can be assigned to no more than one task in given time during the project. The last constraint (Eq. \ref{eq:each_task}) says that each task must be performed in schedule $PS$ by one resource assignment.

\subsection{\textit{Evaluation function}}

As it was mentioned, the proposed approach allows to set various objectives of optimization: duration-- or cost-- oriented one. Those two aspects are normalized, weighted and summarized. Normalization is necessary because of different domains of both aspects that are in opposition to each other. Setting optimization more cost--oriented causes enlarging the project duration, while setting as more important the duration aspect of optimization could increase the cost of the project. 

The detailed formulation of the evaluation function has been presented in Sec. \ref{subsec:eval}.

\subsection{\textit{Solution space size}}
Because of NP--hard (combinatorial) nature of investigated problem, we have decided to present an estimation of solution space size ($SS$). It has been computed as follows:
\begin{equation}\label{eq:space_size}
SS(n,m) = n! * m^{n}
\end{equation}
The above estimation is valid for all solutions, including non--feasible ones. Computing factorial of tasks number provides the number of combinations of ordering tasks within the timeline. It is easy to notice that such estimation allows to set any order, skipping precedence constraints. The second element of Eq. \ref{eq:space_size} provides the number of resource--to--task assignments, including a situation that the same resource is assigned to all tasks and no skill constraints are preserved. 

To imagine how big the solution space could be, let's take into account a sample project schedule with $100$ tasks and $20$ resources. Using Eq. \ref{eq:space_size}, the solution space size is equal to $SS(100,20)=1.19*10^{288}$ solutions, including both feasible and infeasible ones. 


\section{Proposed approach\label{sec:approach}}

Before we describe the details of the proposed approach, some basic ACO definitions in terms of MS--RCPSP should be introduced. \textbf{Colony} is represented as a set of ants: $A={1,...,p}$, where $p$ is a number of ants in population. \textbf{Edge} represents a given task and resources that are capable of performing it. Furthermore, edge stores information about the pheromone ($Ph_j={1,...,p_j^k}$) values for each resource capable of performing a given task. \textbf{Surface} is represented as a set of edges: $E={1,...,j}$ -- all possible task--to--resource assignments, while \textbf{path} represents the set of specified task--to--resource assignments. Path is assigned to a given ant that represents a single solution. Surface represents the solution space of skill--feasible solutions. 

The pheromone value determines the probability of assigning given resource to given task. In the first step of classical ACO, the initial value of pheromone is given for each resource in every edge while for a heuristic initialization, pheromone value is the biggest for path reflecting solution found by heuristic. It means that, at the beginning of our approach run, the probability of choosing resource to be assigned to a task is equal in classical ACO or is close to 1 for path representing heuristic--found solution and close to 0 for remaining edges in the surface. 

Firstly we have used heuristics from \citep{skowronski_heu} to find the best approach for duration optimization (DO) and cost optimization (CO) modes. Based on the obtained results, successors list size--based heuristic (SLS) \citep{skowronski_heu} with descending order has been used for DO and resource salary--based (RS) \citep{skowronski_heu} with ascending order has been used for CO. Output of scheduling project instances by those heuristics has been used as input for ACO method that has been run with the same parameters' configuration as ACO not boosted by heuristic.

The proposed hybrid ACO--based approach could be briefly described in the following steps: 
\begin{enumerate}
\item{Set initial ant population using heuristics to find \textit{good} initial solution}
\item{Check the stopping condition.}
\item{Select edge for each ant.}
\item{Evaluate solutions.}
\item{Evaporate given amount of pheromone from each edge.}
\item{Update solutions.}
\item{Update pheromone value in edges by selected ants.}
\item{Return to 2.}
\end{enumerate}
The pseudocode of investigated HAntCO approach is presented in Alg. \ref{alg:aco}.
\begin{algorithm}[!ht]
\caption{HAntCO pseudocode}
\begin{algorithmic}[1]
\STATE $A \leftarrow set$ \space $initial$ \space $solution$
\WHILE {$stopping$ \space $criterion$ \space $not$ \space $satisfied$}
\FOR {$a \in A$}
\FOR {$e \in Path(a)$}
\STATE $e \leftarrow selectEdge(J^k)$
\STATE $f(a) \leftarrow evaluate(a)$
\ENDFOR
\ENDFOR
\FOR {$e \in E$}
\STATE {$p_e \leftarrow decayPheromone()$}
\ENDFOR 
\STATE $A' \leftarrow selectAnts(A)$
\FOR {$a' \in A'$}
\FOR {$e \in Path(a')$}
\STATE $p_e \leftarrow updatePheromone(e)$
\ENDFOR
\ENDFOR
\STATE   $A_b \leftarrow getBestAnt(A)$
\STATE $A_w \leftarrow getWorstAnt(A)$
\IF {$f(A_b) < f(A_g)$}
\STATE $A_g \leftarrow A_b$
\ENDIF
\IF {$f(A_w) > f(A_v$)}
\STATE $A_v \leftarrow A_w$
\ENDIF
\STATE   $A \leftarrow A'$
\ENDWHILE
\STATE $return$ \space $A_g$
\end{algorithmic}
\label{alg:aco}
\end{algorithm} 

In every iteration, some ants have to be selected (line 9 in Alg. \ref{alg:aco}) to update a pheromone on their edges. The decision which ant should be chosen depends on selected pheromone update methods. There could be all ants chosen, only the local and global best or the local best and worst. Choosing ants to update a pheromone has been described in detail in Subsection \ref{subsec:update}. After each iteration, pheromone values are updated. Then local ($A_b$) and global ($A_g$) best solutions are updated. After each iteration, solutions in ants are ordered ascending by their evaluation function value (line 13). The first ant from the list is set as the best one ($A_b$) while the last one -- as the worst local one. If the evaluation function value of the best local solution ($A_b$) is smaller (minimization problem) than evaluation function value for the best global solution ($A_g$), the best global solution is updated (line 15). Analogously the global worst solution ($A_v$) is updated. The local worst solution ($A_w$) is used in DIFF pheromone update method. 

\subsection{\textit{HAntCO Colony initialization}}
In the first step of classical ACO, the surface of $n$ edges is obtained. For each resource in each edge, the initial pheromone value is set. Then $p$ ants are defined by choosing random capable resource to $j$ task. To reduce the influence of non--determinism and make search more directed, we have decided to introduce a heuristic initialization in hybrid called HAntCO. In HAntCO, one ant has assigned schedule obtained by heuristic described in \citep{skowronski_heu}. This ant is set as favourable -- it can leave much more pheromone than any other ant in a colony. Other ants in the colony are defined in the same way as in classical ACO initial colony definition. 

Heuristic used to obtain an initial solution is varied depending on the optimization mode. For the Duration Optimization mode (DO) the Successors' List Size (SLS) heuristic has been used, as it provided the best results for DO mode. In this method, tasks are sorted by a number of successors they have in ascending order. Then for every task from ordered list a resource is assigned. The decision which resource should be assigned is determined by the earliest time when given resource would finish its previous tasks it has been assigned to. 

For Cost Optimization (CO) mode, resources are sorted ascending by their standard salary rate and then are assigned to tasks from the list given in project definition, preserving skill constraints and avoiding conflicts, by assigning a given task to resource no earlier than all previously assigned tasks to resource would be finished.

In the next step each solution is evaluated, to set the pheromone value for each ant in the next iteration. The amount of pheromone left in every iteration is set according to the ant chosen as the best. 

As the \textbf{stopping criterion}, the number of iterations with no change of global best solution has been proposed in this approach. It is notated as $\gamma$.

The probability of selecting resource $k$ to task $j$ in \textbf{edge selection} bases on the roulette method and is computed as follows:
\begin{equation}
prob_j^k = \frac{{p_j^k}^\alpha}{\sum_{i=1}^n {p_i^k}^\alpha}
\end{equation}
Where $\alpha$ is a weight for pheromone values influence. This value is the parameter of ACO approach and should be provided by the user.  $p^k_j$ is a pheromone value stored in the edge containing information about $k$ resource performing $j$ task.

\subsection{\textit{Evaluation solution method}}\label{subsec:eval}
Evaluation function is formulated as follows:
\begin{equation}\label{eq:eval}
\min{f(PS)} = w_{\tau} f_{\tau}(PS) + (1-w_{\tau}) f_c(PS)
\end{equation}
where: $w_{\tau}$ -- weight of duration component, $f_{\tau}(PS)$ -- duration evaluation component, $f_c(PS)$ -- cost evaluation component. Both components are non--negative values, while $w_{\tau} \in [0;1]$.

Summing both components' weight to 1 ensures that changing the importance of one aspect would cause also some change of second aspect's importance. 

The time component $f_{\tau}(PS)$ is calculated as follows:
\begin{equation}
f_{\tau}(PS)= \frac{\tau}{\tau_{max}}
\end{equation}
where: $\tau_{max}$ -- maximal (pessimistic) possible duration of the schedule $PS$, computed as the sum of all tasks' duration. It occurs when all tasks are performed serially in project: one--by--one. No matter how many and how flexible resources are. 

The cost component $f_c(PS)$ is defined as follows:
\begin{equation}
f_c(PS) = \frac{\sum^J_{i=1} c_j}{c_{max}-c_{min}}
\end{equation}
where: $c_{min}$ -- minimal schedule cost -- the total cost of all tasks assigned to the cheapest resource, $c_{max}$ -- maximal schedule cost -- a total cost of all tasks assigned to the most expensive resource. Note: $c_{max}$ and $c_{min}$ do not involve skill constraints. It means that $c_{min}$ value could be reached only for non--feasible solution. Analogously to $c_{max}$.

\subsection{\textit{Update pheromone}}\label{subsec:update}

Pheromone evaporates iterative. It means the phero-mone value is decreased by the same value ($\mu$) in every iteration, as it was stated in the Eq. \ref{eq:evap}. 
\begin{equation}\label{eq:evap}
(p_j^k)^{(i+1)}=(p_j^k)^{i}(1-\mu)
\end{equation}
Obtained results for various update pheromone methods strongly depend on values set for the following parameters used in ACO:
\begin{itemize}
\item{$p_{init}$ -- initial value of pheromone amount in each edge,}
\item{$\mu$ -- amount of pheromone evaporated in each iteration,}
\item{$\delta$ -- amount of pheromone left in edges by ants, }
\item{$p_{min}$ -- minimal value of pheromone set for resource in edge.}
\end{itemize}
In the proposed approach, three strategies of setting pheromones have been researched: $ALL$ \citep{liang}, $ELITE$ \citep{merkle} and $DIFF$. The last of the proposed ones is the new one, proposed by the authors of this paper. 

\subsubsection{Update pheromone -- ALL}
In this approach, every ant can leave the pheromone value in the edge for selected resource \citep{liang}. The better the solution is, the more pheromone could be left by the ant in given edge. The best ant leaves the pheromone in the amount equal to $\delta$. All next ants leave the amount of pheromone equal to $\delta$ divided by the ant's position ($pos$) in the list ordered ascending by the evaluation function value. 
\begin{equation}
(p_j^k)^{(i+1)} = (p_j^k)^i + \frac{\delta}{pos}
\end{equation}
The main advantage of this approach is the method's resistance to being stuck in local optima. On the other hand, this approach raises a risk of missing the best solutions because of the more exploratory than exploitation--based character of search process. 

\subsubsection{Update pheromone -- ELITE}
In this approach, only elite ants are allowed to leave the pheromone on given edges. The set of elite ants always contains two ants: the one with the best solution found in the current iteration ($A_b$) and the global best one ($A_g$) \citep{merkle} -- with the best solution found from the beginning of search process. For both ants, the same pheromone amount update method is set:
\begin{equation}
(p_j^k)^{(i+1)} = (p_j^k)^i + \delta
\end{equation}
As this approach is more local--optimum oriented, it could lead to getting stuck in local optima. However, the convergence to the optimum of this approach is faster than in  ALL method. 

\subsubsection{Update pheromone -- DIFF}
In this approach, the ant with the worst or best found solution in given iteration is selected. Updating the pheromone value by \textit{the worst} allows to explore the search space in other than potentially the best directions and, consequently, escape from local optima. The same like in ELITE approach, only two ants are able to leave the pheromone: the best ($A_b$) / worst ($A_w$) in iteration and global best ($A_g$) / global worst ($A_v$) found so far. The decision which ant (best or worst) should leave pheromone is made on the basis of satisfaction of the following condition:
\begin{equation}\label{eq:diff_trigger}
\pi > \psi 
\end{equation}
Where $\pi$ is regarded as an ant population variety and is computed as follows:
\begin{equation}\label{eq:pi}
\pi = \frac{f_w-f_b}{f_w}
\end{equation}
Where $f_b$ and $f_w$ are the evaluation function values of the best and worst solutions contained by given ants in specified iteration. The right--sided variable $\psi$ could be regarded as an ant population variety threshold and is set as an ACO parameter. 

If condition in Eq. \ref{eq:diff_trigger} is satisfied, ELITE update pheromone method is used. With every iteration in which condition from Eq. \ref{eq:diff_trigger} is satisfied, the counter ($\kappa$) of possible worst pheromone update strategy use is incremented. If the variety computed in Eq. \ref{eq:diff_trigger} is not satisfied, it means ants are concentrated near some local optima. Then, to avoid being stuck, the worst update method is launched. It means that not the best but worst ants leave pheromone on their path. Meanwhile the counter $\kappa$ is decremented. The worst ant can leave pheromone as long as the ant population variety is smaller than $\psi$ or the $\kappa$ is not negative. Initial $\kappa$ value is also set as an ACO parameter. 

The value of pheromone left by the global ant is defined in Eq. \ref{eq:global}. For the global best (or worst) ant the pheromone amount update value is defined as follows:
\begin{equation}\label{eq:global}
(p_j^k)^{(i+1)} = (p_j^k)^i + \frac{\delta}{\gamma}
\end{equation}
Where $\gamma$ is a number of iterations from the last found new global best. 

For the best / worst ant in iteration, the pheromone amount update value is stated as follows:
\begin{equation}\label{eq:local}
(p_j^k)^{(i+1)} = (p_j^k)^i + \frac{\delta}{\pi}
\end{equation}

In update pheromone amount method for global ant (Eq. \ref{eq:global}) the pheromone amount is reduced, while the pheromone amount for the best local ant is increased (Eq. \ref{eq:local}). It enhances the possibility of finding new global optimum, reducing the probability of losing the best solution found so far at the same time.

\subsection{Conflict fixing \label{subsec:conflict}}

A conflict appears when more than one task is assigned to the same resource in overlapping periods. In that case, it should be fixed by the following procedure. 

It is performed by shifting one of conflicting task's start date. Consequently, the finish date of that task also has to be shifted in order to keep the task duration. The decision which of conflicting tasks should be shifted depends on which of them starts earlier. If they are set to start at exactly the same time, task to be shifted is selected by the way, which was firstly added to project definition. Furthermore, we do not investigate the velocity of resources. Therefore, job duration is constant regardless of assigned resource and skills it owns. 

\begin{figure}[!ht]
\centering
\includegraphics[scale=0.35]{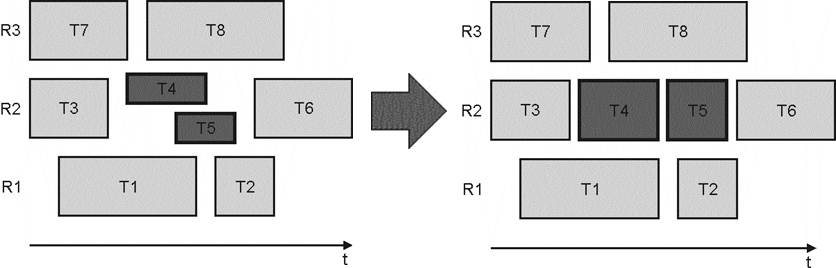}
\caption{Example of conflict resolving}
\label{img:conflict}
\end{figure}

\textbf{Conflict fixing procedure} illustration has been presented in the Fig. \ref{img:conflict}. 

Tasks $T4$ and $T5$ have been assigned to the Resource $R2$ in overlapping period. As a conflict fixing result, a new schedule has been presented, where $T5$ starts just after the $T4$ should be finished. The $T5$ has been shifted, because it was initially set to start later than the $T4$. 
 

\section{Experiments and results\label{sec:experiments}}

The goal of the conducted experiments was to investigate the following issues:
\begin{itemize}
\item{robustness of ACO approach for MS--RCPSP based on given dataset,}
\item{robustness of various update pheromone methods,}
\item{comparing HAntCO to classical ACO approach and other (meta--)heuristics.}
\end{itemize}
Therefore, we have compared the results obtained for different update pheromone methods and results for hybrids and classical ACO approach. Furthermore, the results for simple heuristic scheduling have been provided to get a reference for the ACO--based mechanism. 

The obtained results (project schedules) are described by duration time ([days]) and performance cost ([currency units]). Those project schedule properties have been used to compare the investigated methods. 
\subsection{\textit{iMOPSE} dataset}
Due to evaluating not only the project schedule duration, but also the cost of the schedule, we cannot use the standard PSPLIB benchmark dataset \citep{kolisch} that does not contain any information about the task performance cost. What is more, PSPLIB dataset instances do not reflect the MS--RCPSP. Hence, lack of benchmark data has encouraged us to prepare the \textit{iMOPSE} dataset, containing 36 project instances, that have been artificially created\footnote{http://imopse.ii.pwr.wroc.pl/ -- iMOPSE (intelligent Multi Objective Project Scheduling Environment) project homepage, containing description of investigated methods, dataset definition and best found solutions.}, on the basis of real--world instances, obtained from an international enterprise. We recommend other scientists using \textit{iMOPSE} dataset as a benchmark for investigating their approaches in solving MS--RCPSP as defined. 

Instances of the dataset have been created according to the analysis made in cooperation with experienced project manager from Volvo IT. We were not allowed to get real project data because of their sensitive character for the enterprise. However, we made a statistical analysis of real projects. Then we prepared artificial dataset instances according to the analysis result, regarding the most common project characteristics, like a number of tasks, a number of resources, various skill types in enterprise and the structure of critical chain (a number of tasks involved by precedence relations), etc. 

\begin{table}[!ht]
  \centering
  \caption{iMOPSE dataset summary}
    \small \begin{tabular}{|l|r|r|r|r|r|} \hline
    Dataset instance & Tasks & Resources & Relations & Skills \\ \hline
    100\_20\_23\_9\_D1 & 100   & 20    & 23    & 9 \\
    100\_20\_22\_15 & 100   & 20    & 22    & 15 \\
    100\_20\_47\_9 & 100   & 20    & 47    & 9 \\
    100\_20\_46\_15 & 100   & 20    & 46    & 15 \\
    100\_20\_65\_9 & 100   & 20    & 65    & 9 \\
    100\_20\_65\_15 & 100   & 20    & 65    & 15 \\
    100\_10\_27\_9\_D2 & 100   & 10    & 27    & 9 \\
    100\_10\_26\_15 & 100   & 10    & 26    & 15 \\
    100\_10\_47\_9 & 100   & 10    & 47    & 9 \\
    100\_10\_48\_15 & 100   & 10    & 48    & 15 \\
    100\_10\_64\_9 & 100   & 10    & 64    & 9 \\
    100\_10\_65\_15 & 100   & 10    & 65    & 15 \\
    100\_5\_20\_9\_D3 & 100   & 5     & 20    & 9 \\
    100\_5\_20\_15 & 100   & 5     & 22    & 15 \\
    100\_5\_48\_9 & 100   & 5     & 48    & 9 \\
    100\_5\_48\_15 & 100   & 5     & 46    & 15 \\
    100\_5\_64\_9 & 100   & 5     & 64    & 9 \\
    100\_5\_64\_15 & 100   & 5     & 64    & 15 \\
    200\_40\_45\_9 & 200   & 40    & 45    & 9 \\
    200\_40\_45\_15 & 200   & 40    & 45    & 15 \\
    200\_40\_90\_9 & 200   & 40    & 90    & 9 \\
    200\_40\_91\_9 & 200   & 40    & 91    & 15 \\
    200\_40\_130\_9\_D4 & 200   & 40    & 130   & 9 \\
    200\_40\_144\_15 & 200   & 40    & 133   & 15 \\
    200\_20\_55\_9 & 200   & 20    & 55    & 9 \\
    200\_20\_54\_15 & 200   & 20    & 54    & 15 \\
    200\_20\_97\_9 & 200   & 20    & 97    & 9 \\
    200\_20\_97\_15 & 200   & 20    & 97    & 15 \\
    200\_20\_150\_9\_D5 & 200   & 20    & 150   & 9 \\
    200\_20\_145\_15 & 200   & 20    & 145   & 15 \\
    200\_10\_50\_9 & 200   & 10    & 50    & 9 \\
    200\_10\_50\_15 & 200   & 10    & 50    & 15 \\
    200\_10\_84\_9 & 200   & 10    & 84    & 9 \\
    200\_10\_85\_15 & 200   & 10    & 85    & 15 \\
    200\_10\_135\_9\_D6 & 200   & 10    & 135   & 9 \\
    200\_10\_128\_15 & 200   & 10    & 128   & 15 \\ \hline
    \end{tabular}%
  \label{tab:dataset_summary}%
\end{table}%

The iMOPSE dataset summary has been presented in the Table \ref{tab:dataset_summary}. There are two groups of created project instances: one contains 100 tasks and the second -- 200 tasks. Within each group, project instances are varied by a number of available resources and the precedence relationship complexity. The number of resources for instances from both groups were chosen in a way to preserve constant average resource load and average task relations ratio for given instances. Hence for project instances with 200 tasks the number of possible resources and precedence relations is twice bigger than for project instances containing 100 tasks. The skill variety has been set up to 9 or 15 different skill types for each project instance, while any resource can dispose of exactly six different skill types. Because of the different resources and relations number, the scheduling complexity for each project is varied. 

This dataset stands as an extension of dataset presented in \citep{skowronski_men}, \citep{skowronski_ts}, \citep{skowronski_heu} and that is the reason some instances are named with suffix Dx. This suffix refers to dataset instances that have been previously created and presented in those papers. Because of the extension of the dataset, the need of introducing more clear namesystem has arisen. Suffix has been added to a reference of previously created files, keeping the naming convention applied after dataset extension. 

\subsection{Experiments' set--up}

The experiments have been divided into investigating the influence of ACO parameters' configurations for project duration and performance cost in three various components' weights in evaluation function: duration optimization (DO: $w_\tau=1$, see. Eq. \ref{eq:eval}), balanced optimization (BO: $w_\tau=0.5$) and cost optimization (CO: $w_\tau=0$). Because of the stochastic nature of ACO--based methods, each experiment for given parameter configuration has been repeated ten times. For K--S test and t--test each experiment has been repeated 50 times (see Tab.\ref{tab:CDComparison} and Tab.\ref{tab:STATScomparision}). On the other hand, deterministic character of heuristics allowed us to obtain results for those methods in only one iteration for every parameters' configuration.

The further step of the conducted experiments was to compare results obtained for random initial solution with boosting initial solution by using described hybrids. Initial solution has been previously obtained by using the above--mentioned heuristics and then set them as input for ACO and made those results as more favourable in local search by enhancing the pheromone value left in this path representing initial solution. We decided to use SLS(D) \citep{skowronski_heu} for DO mode and RS(A) \citep{skowronski_heu} for CO mode optimization within HAntCO hybrid. Because of some code refactoring, we were able to tune our heuristics and obtain a better solution than the found ones in \citep{skowronski_heu}. That is the reason why the results of those heuristics in this paper are slightly better than the results in \citep{skowronski_heu} for given dataset instances. 

\begin{table*}[!ht]
  \centering
  \caption{Comparison of the best obtained results for DO, BO and CO modes in classical ACO and selected heuristics from \citep{skowronski_heu}}
    \tiny \begin{tabular}{|r|r|r|r|r|r|r|r|r|r|r|r|r|r|r|} \hline
    \multicolumn{1}|{|c|}{\multirow{3}[0]{*}{Dataset instance}} & \multicolumn{9}{|c|}{ACO}    & \multicolumn{5}{|c|}{Heuristics}   \\ \cline{2-15}
    \multicolumn{1}{|c|}{} & \multicolumn{3}{|c|}{DO} & \multicolumn{3}{|c|}{BO} & \multicolumn{3}{c}{CO} & \multicolumn{2}{|c|}{DO} & \multicolumn{3}{|c|}{CO} \\ \cline{2-15}
    \multicolumn{1}{|c|}{} & M     & Days  & Cost  & M     & Days  & Cost  & M     & Days  & Cost  & Days  & Cost  & C     & Days  & Cost \\ \hline
    100\_10\_26\_15 & E     & \textbf{32} & 124687 & E/D   & 85    & 70326 & E/D   & 85    & \textbf{70326} & 37    & 126361 & RS(A) & 85    & \textbf{70326} \\
    100\_10\_27\_9\_D2 & E     & \textbf{34} & 44999 & D     & 72    & 27120 & E/D   & 129   & \textbf{26323} & 38    & 44309 & RS(A) & 129   & \textbf{26323} \\
    100\_10\_47\_9 & E     & \textbf{36} & 143100 & D     & 105   & 94334 & E/D   & 145   & \textbf{90992} & 41    & 142759 & RS(A) & 145   & \textbf{90992} \\
    100\_10\_48\_15 & E     & \textbf{33} & 133062 & E/D   & 81    & 87194 & E     & 85    & \textbf{87187} & 36    & 135534 & RS(A) & 85    & \textbf{87187} \\
    100\_10\_64\_9 & D     & \textbf{35} & 110643 & D     & 92    & 63934 & E/D   & 121   & \textbf{62102} & 39    & 113124 & RS(A) & 121   & \textbf{62102} \\
    100\_10\_65\_15 & E     & \textbf{35} & 150294 & E/D   & 76    & 108312 & E/D   & 98    & \textbf{106296} & 40    & 152955 & RS(A) & 98    & \textbf{106296} \\
    100\_20\_22\_15 & D     & \textbf{20} & 120949 & D     & 56    & 56625 & D     & 87    & 55240 & 25    & 117493 & ADAD  & 86    & \textbf{55240} \\
    100\_20\_23\_9\_D1 & D     & \textbf{32} & 52119 & D     & 60    & 30900 & D     & 121   & 30107 & 32    & 53154 & AAAD  & 119   & \textbf{30104} \\
    100\_20\_46\_15 & E     & \textbf{25} & 138565 & D     & 65    & 69789 & E/D   & 75    & \textbf{68899} & 28    & 138270 & RS(A) & 75    & \textbf{68899} \\
    100\_20\_47\_9 & E     & \textbf{21} & 124817 & D     & 69    & 59196 & D     & 131   & \textbf{55197} & 21    & 129160 & RS(A) & 131   & \textbf{55197} \\
    100\_20\_65\_15 & E     & \textbf{27} & 109831 & D     & 52    & 57338 & E/D   & 69    & \textbf{57085} & 32    & 110503 & RS(A) & 69   & 57085 \\
    100\_20\_65\_9 & E     & \textbf{23} & 130934 & D     & 76    & 61913 & D     & 114   & 59736 & 25    & 127149 & RS(A) & 114    & \textbf{59736} \\
    100\_5\_20\_9\_D3 & E     & \textbf{50} & 41029 & D     & 75    & 31681 & E/D   & 167   & \textbf{30164} & 57    & 40539 & RS(A) & 167   & \textbf{30164} \\
    100\_5\_22\_15 & D     & \textbf{60} & 119434 & D     & 70    & 110145 & E/D   & 86    & \textbf{109111} & 63    & 119266 & RS(A) & 86    & \textbf{109111} \\
    100\_5\_46\_15 & E     & \textbf{67} & 204110 &  *    & 125   & 184409 & E/D   & 125   & \textbf{184409} & 75    & 202238 & RS(A) & 125   & \textbf{184409} \\
    100\_5\_48\_9 & E     & \textbf{62} & 191712 & E/D   & 127   & 175526 & E/D   & 130   & \textbf{175225} & 72    & 193383 & RS(A) & 130   & \textbf{175225} \\
    100\_5\_64\_15 & D     & \textbf{62} & 144972 & E/D   & 123   & 109431 & E/D   & 141   & \textbf{109091} & 71    & 141407 & RS(A) & 141   & \textbf{109091} \\
    100\_5\_64\_9 & E     & \textbf{61} & 102777 & D     & 87    & 74617 & E/D   & 173   & \textbf{72848} & 71    & 102439 & RS(A) & 173   & \textbf{72848} \\
    200\_10\_128\_15 & E     & \textbf{62} & 178264 & D     & 126   & 136643 & E     & 143   & 136551 & 71    & 180812 & AxAD  & 159   & \textbf{134425} \\
    200\_10\_135\_9\_D6 &  *    & \textbf{216} & 99375 & E     & 237   & 72753 & D     & 274   & 72036 & 216   & 105593 & RS(A) & 256   & \textbf{71986} \\
    200\_10\_50\_15 & E     & \textbf{63} & 191856 & D     & 144   & 85712 & E/D   & 167   & \textbf{84308} & 66    & 189660 & RS(A) & 167   & \textbf{84308} \\
    200\_10\_50\_9 & E     & \textbf{65} & 250075 & D     & 228   & 110218 & D     & 318   & 105232 & 66    & 251158 & RS(A) & 318   & \textbf{105198} \\
    200\_10\_84\_9 & E     & \textbf{69} & 226666 & D     & 171   & 125715 & D     & 316   & 117754 & 70    & 224121 & DAAA  & 338   & 117543 \\
    200\_10\_85\_15 & E     & \textbf{61} & 306949 & E     & 180   & 197767 & E     & 215   & \textbf{195820} & 65    & 304277 & RS(A) & 215   & \textbf{195820} \\
    200\_20\_145\_15 & E     & 36    & 278199 & D     & 109   & 144694 & D     & 152   & 143688 & \textbf{36} & 275983 & RS(A) & 158   & \textbf{143497} \\
    200\_20\_150\_9\_D5 & D     & 186   & 91461 & D     & 247   & 52620 & D     & 296   & 51678 & \textbf{183} & 92821 & ADDA  & 337   & \textbf{51496} \\
    200\_20\_54\_15 & E     & 39    & 299993 & D     & 123   & 161883 & D     & 131   & 161614 & \textbf{37} & 295786 & RS(A) & 125   & \textbf{161412} \\
    200\_20\_55\_9 & D     & 38    & 231094 & D     & 159   & 75836 & D     & 250   & 72176 & \textbf{37} & 230150 & RS(A) & 332   & \textbf{70057} \\
    200\_20\_97\_15 & D     & \textbf{42} & 280951 & D     & 115   & 160070 & D     & 169   & 157202 & 49    & 290399 & RS(A) & 171   & \textbf{156951} \\
    200\_20\_97\_9 & E     & 37    & 275819 & D     & 114   & 102641 & D     & 150   & 99901 & \textbf{35} & 273378 & RS(A) & 169   & \textbf{98480} \\
    200\_40\_130\_9\_D4 &  *    & \textbf{112} & 94488 & D     & 132   & 48362 & D     & 205   & 48419 & 112   & 101879 & DAAD  & 214   & \textbf{46133} \\
    200\_40\_133\_15 & D     & 27    & 281933 & D     & 93    & 101620 & D     & 131   & 99329 & \textbf{24} & 276456 & AAAA  & 155   & \textbf{97345} \\
    200\_40\_45\_15 & E     & \textbf{25} & 248717 & D     & 118   & 95959 & D     & 161   & 91010 & 31    & 260738 & RS(A) & 213   & \textbf{87955} \\
    200\_40\_45\_9 & E     & 26    & 273632 & D     & 118   & 96375 & D     & 179   & 94142 & \textbf{22} & 270758 & AAAA  & 334   & \textbf{77236} \\
    200\_40\_90\_9 & E     & 26    & 287694 & D     & 115   & 97926 & D     & 142   & 96312 & \textbf{24} & 290028 & RS(A) & 285   & \textbf{80732} \\
    200\_40\_91\_15 & E     & 25    & 257927 & D     & 82    & 91204 & D     & 132   & 88616 & \textbf{19} & 249909 & RS(A) & 184   & \textbf{86476} \\ \hline
    \end{tabular}%
  \label{tab:best}%
\end{table*}%

To present averaged results in detail (see Tab. \ref{tab:average}), a standard deviation measure ($\sigma$) has been introduced and applied to each average value, presented as a percentage value in relation to the average. We have also added information about the best found solution for a given method (see Tab. \ref{tab:best}) that have been compared with the results obtained by most promising heuristics, described in \citep{skowronski_heu}. 

Both for the best and averaged results, pheromone update methods have been compared and the one that provided best results (shortest duration in DO, smallest cost in CO and smallest evaluation function value in BO) has been presented in Tab. \ref{tab:average} and Tab. \ref{tab:best}. The notation for methods used in tables with obtained results is as the following: E - update ELITE pheromone method, A - update ALL, D - update DIFF. If more than one pheromone update methods turned out to be the best and gave the same results, they have been presented both separated by "$\slash$" (e.g.: E$\slash$D -- both update DIFF and update ELITE methods gave the same, best results). In Tab. \ref{tab:best} a sign $*$ has been also introduced to indicate a situation where all three methods provided the same, regarded as the best, result.

\begin{table*}[!ht]
  \centering
  \caption{Best results obtained for HAntCO with various pheromone update methods in DO and CO optimization mode}
    \tiny \begin{tabular}{|r|r|r|r|r|r|r|r|r|} \hline
    \multicolumn{1}{|c|}{\multirow{3}[0]{*}{Dataset instance}} & \multicolumn{4}{|c|}{DO}        & \multicolumn{4}{|c|}{CO}        \\ \cline{2-9}
    \multicolumn{1}{|c|}{} & \multicolumn{2}{|c|}{ELITE} & \multicolumn{2}{|c|}{DIFF} & \multicolumn{2}{|c|}{ELITE} & \multicolumn{2}{|c|}{DIFF} \\ \cline{2-9}
    \multicolumn{1}{|c|}{} & \multicolumn{1}{|c|}{days} & \multicolumn{1}{|c|}{cost} & \multicolumn{1}{|c|}{days} & \multicolumn{1}{|c|}{cost} & \multicolumn{1}{|c|}{days} & \multicolumn{1}{|c|}{cost} & \multicolumn{1}{|c|}{days} & \multicolumn{1}{|c|}{cost} \\ \hline
    100\_10\_26\_15 & \textbf{31} & 126216 & 32    & 125688 & 85    & 70326 & 85    & 70326 \\
    100\_10\_27\_9\_D2 & \textbf{33} & 42199 & 35    & 44022 & 129   & 26323 & 129   & 26323 \\
    100\_10\_47\_9 & \textbf{34} & 140865 & 34    & 142362 & 145   & 90992 & 145   & 90992 \\
    100\_10\_48\_15 & 33    & 134692 & \textbf{33} & 133495 & 85    & 87187 & 85    & 87187 \\
    100\_10\_64\_9 & \textbf{33} & 113774 & 34    & 115998 & 121   & 62102 & 121   & 62102 \\
    100\_10\_65\_15 & 33    & 149175 & \textbf{32} & 149185 & 98    & 106296 & 98    & 106296 \\
    100\_20\_22\_15 & \textbf{19} & 123642 & 20    & 118054 & 87    & 55240 & 87    & 55240 \\
    100\_20\_23\_9\_D1 & \textbf{23} & 53358 & 24    & 54309 & 117   & 30104 & 117   & 30104 \\
    100\_20\_46\_15 & \textbf{24} & 138568 & 24    & 142206 & 75    & 68899 & 75    & 68899 \\
    100\_20\_47\_9 & \textbf{18} & 134312 & 21    & 133050 & 131   & 55197 & 131   & 55197 \\
    100\_20\_65\_15 & \textbf{27} & 108991 & 27    & 113275 & 69    & 57085 & 69    & 57085 \\
    100\_20\_65\_9 & 21    & 126659 & \textbf{20} & 128354 & 114   & 59736 & 114   & 59736 \\
    100\_5\_20\_9\_D3 & 53    & 41310 & \textbf{53} & 40811 & 167   & 30164 & 167   & 30164 \\
    100\_5\_22\_15 & \textbf{60} & 119158 & 61    & 119218 & 86    & 109111 & 86    & 109111 \\
    100\_5\_46\_15 & \textbf{67} & 204730 & 70    & 205618 & 125   & 184409 & 125   & 184409 \\
    100\_5\_48\_9 & \textbf{62} & 191888 & 62    & 192315 & 130   & 175225 & 130   & 175225 \\
    100\_5\_64\_15 & 61    & 145322 & \textbf{61} & 143956 & 141   & 109091 & 141   & 109091 \\
    100\_5\_64\_9 & \textbf{61} & 101297 & 62    & 103777 & 173   & 72848 & 173   & 72848 \\
    200\_10\_128\_15 & \textbf{60} & 178375 & 61    & 180400 & 143   & 136551 & 143   & 136551 \\
    200\_10\_135\_9\_D6 & \textbf{186} & 103561 & 186   & 105515 & 269   & \textbf{71986} & 270   & 71986 \\
    200\_10\_50\_15 & \textbf{62} & 190956 & 62    & 191149 & 167   & 84308 & 167   & 84308 \\
    200\_10\_50\_9 & \textbf{63} & 253214 & 64    & 250850 & 318   & 105198 & 318   & 105198 \\
    200\_10\_84\_9 & 67    & 224639 & \textbf{66} & 222655 & 318   & 117543 & 318   & 117543 \\
    200\_10\_85\_15 & 62    & 303301 & \textbf{62} & 302064 & 215   & 195820 & 215   & 195820 \\
    200\_20\_145\_15 & \textbf{35} & 272504 & 35    & 277291 & 158   & 143497 & 158   & 143497 \\
    200\_20\_150\_9\_D5 & 187   & 90548 & \textbf{177} & 92567 & 344   & 51524 & 345   & \textbf{51496} \\
    200\_20\_54\_15 & \textbf{34} & 298822 & 36    & 295819 & 125   & 161412 & 125   & 161412 \\
    200\_20\_55\_9 & \textbf{36} & 223879 & 36    & 227449 & 311   & 70967 & 332   & \textbf{70057} \\
    200\_20\_97\_15 & 42    & 290308 & \textbf{42} & 277860 & 171   & 156951 & 171   & 156951 \\
    200\_20\_97\_9 & \textbf{35} & 278797 & 36    & 270910 & 155   & 99190 & 169   & \textbf{98480} \\
    200\_40\_130\_9\_D4 & 108   & 106637 & \textbf{108} & 104965 & 225   & 47212 & 216   & \textbf{46275} \\
    200\_40\_133\_15 & \textbf{24} & 282730 & 24    & 279073 & 141   & 97953 & 144   & \textbf{97345} \\
    200\_40\_45\_15 & \textbf{23} & 256687 & 23    & 256753 & 201   & 89407 & 213   & \textbf{87955} \\
    200\_40\_45\_9 & \textbf{25} & 270428 & 26    & 263162 & 270   & 89123 & 315   & \textbf{82192} \\
    200\_40\_90\_9 & \textbf{24} & 298340 & 25    & 293098 & 229   & 93090 & 247   & \textbf{84038} \\
    200\_40\_91\_15 & \textbf{23} & 241492 & 23    & 248984 & 176   & 87875 & 184   & \textbf{86476} \\ \hline
    \end{tabular}%
  \label{tab:hantco_best}%
\end{table*}%

\begin{table*}[!ht]
  \centering
  \caption{Averaged results obtained for classical ACO in various optimization modes}
    \tiny \begin{tabular}{|r|r|r|r|r|r|r|r|r|r|r|r|r|r|r|r|} \hline
    \multicolumn{1}{|c|}{\multirow{3}[0]{*}{Dataset instance}} & \multicolumn{5}{c}{DO} & \multicolumn{5}{|c|}{BO} & \multicolumn{5}{|c|}{CO} \\ \cline{2-16}
    \multicolumn{1}{|c|}{} & \multicolumn{1}{c}{\multirow{2}[0]{*}{M}} & \multicolumn{2}{|c|}{Days} & \multicolumn{2}{|c|}{Cost} & \multicolumn{1}{c}{\multirow{2}[0]{*}{M}} & \multicolumn{2}{|c|}{Days} & \multicolumn{2}{|c|}{Cost} & \multicolumn{1}{|c|}{\multirow{2}[0]{*}{M}} & \multicolumn{2}{|c|}{Days} & \multicolumn{2}{|c|}{Cost} \\ \cline{3-6} \cline{8-11} \cline{13-16}
    \multicolumn{1}{|c|}{} & \multicolumn{1}{|c|}{} & Avg   & $\sigma$   & Avg   & $\sigma$   & \multicolumn{1}{|c|}{} & Avg   & $\sigma$   & Avg   & $\sigma$   & \multicolumn{1}{|c|}{} & Avg   & $\sigma$   & Avg   & $\sigma$ \\ \hline
    100\_10\_26\_15 & E     & 33.2  & 2.6   & 125436 & 1.5   & D     & 85    & 0.0   & 70326 & 0.0   & E     & 84.9  & 0.4   & 70363 & 0.1 \\
    100\_10\_27\_9\_D2 & E     & 36.2  & 4.1   & 43382 & 1.8   & E     & 75.4  & 2.1   & 27064 & 0.1   & E     & 130.5 & 3.5   & 26326 & 0.0 \\
    100\_10\_47\_9 & E     & 37.5  & 2.7   & 142742 & 0.4   & D     & 104.9 & 1.3   & 94501 & 0.3   & D     & 144.8 & 0.4   & 91088 & 0.1 \\
    100\_10\_48\_15 & E     & 35.2  & 4.0   & 135563 & 2.0   & E     & 81    & 0.0   & 87214 & 0.0   & D     & 85.3  & 0.5   & 87205 & 0.0 \\
    100\_10\_64\_9 & E     & 36.8  & 2.7   & 114538 & 1.8   & D     & 90.5  & 1.3   & 64231 & 0.4   & D     & 121   & 0.0   & 62136 & 0.1 \\
    100\_10\_65\_15 & E     & 35.8  & 3.0   & 152033 & 1.2   & E     & 76.7  & 1.4   & 108266 & 0.1   & E     & 98    & 0.0   & 106299 & 0.0 \\
    100\_20\_22\_15 & D     & 22    & 4.5   & 118254 & 2.9   & E     & 52.5  & 3.8   & 57503 & 1.0   & E     & 84.5  & 4.0   & 55431 & 0.3 \\
    100\_20\_23\_9\_D1 & A     & 32    & 0.0   & 52915 & 2.5   & D     & 63.2  & 3.2   & 31009 & 0.5   & D     & 115.7 & 9.2   & 30212 & 0.7 \\
    100\_20\_46\_15 & E     & 24.9  & 3.3   & 140271 & 2.4   & D     & 67.4  & 3.6   & 69574 & 0.4   & D     & 75.2  & 0.8   & 68932 & 0.1 \\
    100\_20\_47\_9 & E     & 23.3  & 5.1   & 128127 & 3.2   & D     & 69.7  & 3.1   & 59802 & 0.9   & E     & 116.6 & 7.8   & 56800 & 1.8 \\
    100\_20\_65\_15 & E     & 27.2  & 1.5   & 111946 & 4.0   & E     & 51.4  & 2.3   & 57645 & 0.5   & E     & 66.9  & 3.7   & 57131 & 0.1 \\
    100\_20\_65\_9 & E     & 23.9  & 2.3   & 126709 & 2.8   & E     & 71.5  & 4.5   & 64189 & 2.7   & D     & 103.1 & 10.5  & 60929 & 2.6 \\
    100\_5\_20\_9\_D3 & E     & 52.4  & 2.4   & 41152 & 1.1   & E     & 76.5  & 2.0   & 31653 & 0.1   & E     & 166.9 & 0.2   & 30167 & 0.0 \\
    100\_5\_22\_15 & E     & 61    & 0.7   & 119479 & 0.4   & E     & 70.2  & 0.6   & 110135 & 0.0   & E     & 86    & 0.0   & 109111 & 0.0 \\
    100\_5\_46\_15 & E     & 68.2  & 1.7   & 204507 & 0.3   & E     & 125   & 0.0   & 184409 & 0.0   & E     & 125   & 0.0   & 184409 & 0.0 \\
    100\_5\_48\_9 & E     & 63.1  & 1.1   & 191911 & 0.2   & E     & 127   & 0.0   & 175535 & 0.0   & E     & 130   & 0.0   & 175225 & 0.0 \\
    100\_5\_64\_15 & E     & 62.6  & 0.8   & 144257 & 0.7   & D     & 123.1 & 0.2   & 109428 & 0.0   & /   & 141   & 0.0   & 109091 & 0.0 \\
    100\_5\_64\_9 & E     & 63    & 1.9   & 103527 & 1.3   & D     & 87    & 0.0   & 74617 & 0.0   & E     & 172.9 & 0.2   & 72850 & 0.0 \\
    200\_10\_128\_15 & E     & 63.3  & 1.9   & 178421 & 1.2   & D     & 124.9 & 1.1   & 136938 & 0.2   & E     & 140.7 & 1.3   & 136568 & 0.0 \\
    200\_10\_135\_9\_D6 & E     & 216   & 0.0   & 100758 & 1.6   & D     & 247.2 & 1.8   & 72693 & 0.5   & D     & 267.3 & 1.2   & 72127 & 0.1 \\
    200\_10\_50\_15 & E     & 65.3  & 1.9   & 190271 & 2.2   & E     & 134.3 & 3.2   & 87158 & 0.6   & E     & 166.7 & 0.4   & 84402 & 0.1 \\
    200\_10\_50\_9 & E     & 66.6  & 1.8   & 247741 & 1.7   & E     & 220.5 & 2.8   & 113340 & 1.6   & D     & 311   & 3.0   & 105825 & 0.8 \\
    200\_10\_84\_9 & E     & 71.1  & 2.0   & 224680 & 1.9   & E     & 162.1 & 2.0   & 129065 & 1.2   & E     & 275.7 & 7.2   & 121478 & 1.3 \\
    200\_10\_85\_15 & E     & 64.3  & 2.2   & 307437 & 1.0   & E     & 170.2 & 3.6   & 199332 & 0.7   & D     & 212.3 & 1.7   & 196662 & 0.4 \\
    200\_20\_145\_15 & E     & 38.3  & 2.6   & 272720 & 1.8   & D     & 108.3 & 2.1   & 146285 & 0.9   & D     & 143.2 & 10.5  & 144947 & 1.1 \\
    200\_20\_150\_9\_D5 & D     & 190.7 & 1.3   & 91095 & 3.2   & D     & 237   & 3.1   & 54032 & 2.3   & D     & 266.9 & 12.1  & 54512 & 8.3 \\
    200\_20\_54\_15 & E     & 41.2  & 3.4   & 288063 & 2.2   & D     & 124.3 & 1.7   & 162514 & 0.4   & D     & 133.3 & 4.4   & 162498 & 0.4 \\
    200\_20\_55\_9 & D     & 39.7  & 1.6   & 228459 & 2.5   & D     & 148.3 & 9.5   & 80793 & 8.5   & D     & 230.5 & 8.3   & 75247 & 4.3 \\
    200\_20\_97\_15 & D     & 43.3  & 2.7   & 287731 & 1.6   & D     & 114.9 & 2.6   & 160892 & 0.4   & D     & 160.5 & 11.1  & 158560 & 1.6 \\
    200\_20\_97\_9 & D     & 40.8  & 3.3   & 281754 & 2.0   & D     & 112.3 & 2.7   & 105641 & 2.9   & D     & 134   & 5.2   & 101992 & 1.7 \\
    200\_40\_130\_9\_D4 & E     & 112   & 0.0   & 102221 & 3.4   & D     & 141.7 & 9.8   & 51413 & 11.9  & D     & 185.1 & 7.2   & 49156 & 1.6 \\
    200\_40\_133\_15 & E     & 28.4  & 3.2   & 282463 & 2.2   & D     & 89.7  & 3.1   & 104442 & 1.9   & D     & 116.5 & 10.6  & 102689 & 3.0 \\
    200\_40\_45\_15 & E     & 26.9  & 3.9   & 247230 & 3.8   & D     & 106.8 & 7.2   & 102650 & 4.0   & D     & 160.8 & 9.8   & 94330 & 3.6 \\
    200\_40\_45\_9 & E     & 28.2  & 4.1   & 267910 & 2.1   & D     & 102.6 & 10.4  & 106705 & 6.6   & D     & 182.8 & 8.3   & 97018 & 2.0 \\
    200\_40\_90\_9 & E     & 27.4  & 3.3   & 288861 & 2.0   & D     & 109.3 & 12.1  & 104403 & 8.2   & D     & 133   & 13.1  & 102871 & 7.3 \\
    200\_40\_91\_15 & E     & 26.4  & 3.5   & 242588 & 2.4   & D     & 80.2  & 7.9   & 96756 & 6.8   & D     & 112.2 & 10.8  & 92724 & 4.0 \\ \hline
    \end{tabular}%
  \label{tab:average}%
\end{table*}%

\begin{table*}[!ht]
  \centering
  \caption{Averaged results obtained for HAntCO with various pheromone update methods in DO and CO optimization mode}
   \tiny \begin{tabular}{|r|r|r|r|r|r|r|r|r|r|r|r|r|r|r|r|r|} \hline
    \multicolumn{1}{|c|}{\multirow{4}[0]{*}{Dataset instance}} & \multicolumn{8}{|c|}{DO} & \multicolumn{8}{|c|}{CO}   \\ \cline{2-17}
    \multicolumn{1}{|c|}{} & \multicolumn{4}{|c|}{ELITE}     & \multicolumn{4}{|c|}{DIFF}      & \multicolumn{4}{|c|}{ELITE}     & \multicolumn{4}{|c|}{DIFF}      \\ \cline{2-17}
    \multicolumn{1}{|c|}{} & \multicolumn{2}{|c|}{Days} & \multicolumn{2}{|c|}{Cost} & \multicolumn{2}{|c|}{Days} & \multicolumn{2}{|c|}{Cost} & \multicolumn{2}{|c|}{Days} & \multicolumn{2}{|c|}{Cost} & \multicolumn{2}{|c|}{Days} & \multicolumn{2}{|c|}{Cost} \\ \cline{2-17}
    \multicolumn{1}{|c|}{} & \multicolumn{1}{|c|}{Avg} & \multicolumn{1}{|c|}{$\sigma$} & \multicolumn{1}{|c|}{Avg} & \multicolumn{1}{|c|}{$\sigma$} & \multicolumn{1}{|c|}{Avg} & \multicolumn{1}{||c}{$\sigma$} & \multicolumn{1}{|c|}{Avg} & \multicolumn{1}{|c|}{$\sigma$} & \multicolumn{1}{|c|}{Avg} & \multicolumn{1}{|c|}{$\sigma$} & \multicolumn{1}{|c|}{Avg} & \multicolumn{1}{|c|}{$\sigma$} & \multicolumn{1}{|c|}{Avg} & \multicolumn{1}{|c|}{$\sigma$} & \multicolumn{1}{|c|}{Avg} & \multicolumn{1}{|c|}{$\sigma$} \\ \hline
    100\_10\_26\_15 & \textbf{32.5} & 0.92  & 125889 & 1498  & 32.6  & 0.49  & 125848 & 1373  & 85    & 0.00  & 70326 & 0     & 85    & 0.00  & 70326 & 0 \\
    100\_10\_27\_9\_D2 & \textbf{35.1} & 1.37  & 43644 & 661   & 35.8  & 0.75  & 43992 & 650   & 129   & 0.00  & 26323 & 0     & 129   & 0.00  & 26323 & 0 \\
    100\_10\_47\_9 & \textbf{34.9} & 1.04  & 142103 & 998   & 35.2  & 0.75  & 143263 & 944   & 145   & 0.00  & 90992 & 0     & 145   & 0.00  & 90992 & 0 \\
    100\_10\_48\_15 & \textbf{34} & 0.63  & 134504 & 1507  & 34.4  & 0.66  & 134568 & 1509  & 85    & 0.00  & 87187 & 0     & 85    & 0.00  & 87187 & 0 \\
    100\_10\_64\_9 & \textbf{34.7} & 1.10  & 113638 & 1871  & 34.9  & 0.54  & 113230 & 1899  & 121   & 0.00  & 62102 & 0     & 121   & 0.00  & 62102 & 0 \\
    100\_10\_65\_15 & 33.6  & 0.66  & 149474 & 963   & \textbf{33.2} & 0.60  & 149598 & 1033  & 98    & 0.00  & 106296 & 0     & 98    & 0.00  & 106296 & 0 \\
    100\_20\_22\_15 & 20.7  & 1.00  & 118914 & 2464  & \textbf{20.6} & 0.49  & 118347 & 2895  & 87    & 0.00  & 55240 & 0     & 87    & 0.00  & 55240 & 0 \\
    100\_20\_23\_9\_D1 & \textbf{24.5} & 0.81  & 53810 & 1028  & 25    & 0.77  & 53051 & 1243  & 117   & 0.00  & 30104 & 0     & 117   & 0.00  & 30104 & 0 \\
    100\_20\_46\_15 & \textbf{24.2}  & 0.40  & 140491 & 2823  & 24.2  & 0.40  & 141045 & 3922  & 75    & 0.00  & 68899 & 0     & 75    & 0.00  & 68899 & 0 \\
    100\_20\_47\_9 & \textbf{20.3} & 1.10  & 128641 & 2938  & 21.7  & 0.46  & 127577 & 3023  & 131   & 0.00  & 55204 & 19    & 131   & 0.00  & \textbf{55197} & 0 \\
    100\_20\_65\_15 & 27.2  & 0.40  & 111842 & 2758  & \textbf{27} & 0.00  & 113219 & 2501  & 69    & 0.00  & 57085 & 0     & 69    & 0.00  & 57085 & 0 \\
    100\_20\_65\_9 & 21.9  & 0.70  & 126081 & 1789  & \textbf{21.6} & 0.80  & 125269 & 4271  & 114   & 0.90  & 59744 & 24    & 114   & 0.00  & \textbf{59736} & 0 \\
    100\_5\_20\_9\_D3 & \textbf{53.3} & 0.46  & 40917 & 238   & 54.4  & 0.80  & 41025 & 148   & 167   & 0.00  & 30164 & 0     & 167   & 0.00  & 30164 & 0 \\
    100\_5\_22\_15 & \textbf{61.4} & 0.80  & 119219 & 486   & 61.9  & 0.83  & 118934 & 787   & 86    & 0.00  & 109111 & 0     & 86    & 0.00  & 109111 & 0 \\
    100\_5\_46\_15 & \textbf{69.8} & 1.54  & 205451 & 555   & 70.9  & 0.30  & 204973 & 615   & 125   & 0.00  & 184409 & 0     & 125   & 0.00  & 184409 & 0 \\
    100\_5\_48\_9 & \textbf{62.8} & 0.40  & 191934 & 171   & 63    & 0.45  & 192103 & 342   & 130   & 0.00  & 175225 & 0     & 130   & 0.00  & 175225 & 0 \\
    100\_5\_64\_15 & \textbf{62.6} & 1.02  & 144256 & 1342  & 62.9  & 0.94  & 144077 & 813   & 141   & 0.00  & 109091 & 0     & 141   & 0.00  & 109091 & 0 \\
    100\_5\_64\_9 & \textbf{62.5} & 1.12  & 102901 & 1226  & 62.8  & 0.75  & 103495 & 751   & 173   & 0.00  & 72848 & 0     & 173   & 0.00  & 72848 & 0 \\
    200\_10\_128\_15 & \textbf{61.1} & 1.14  & 179159 & 1773  & 61.8  & 0.40  & 178981 & 1685  & 143   & 0.00  & 136551 & 0     & 143   & 0.00  & 136551 & 0 \\
    200\_10\_135\_9\_D6 & 190.9 & 7.53  & 103411 & 2442  & \textbf{186.8} & 2.40  & 104042 & 2117  & 268   & 2.69  & 71986 & 0     & 268.7 & 1.73  & 71986 & 0 \\
    200\_10\_50\_15 & \textbf{63.4} & 1.43  & 188265 & 2814  & 63.8  & 1.08  & 189963 & 2903  & 167   & 0.00  & 84308 & 0     & 167   & 0.00  & 84308 & 0 \\
    200\_10\_50\_9 & \textbf{64} & 0.77  & 250681 & 2505  & 64.8  & 0.40  & 249281 & 1911  & 318   & 0.00  & \textbf{105198} & 1     & 317.6 & 1.20  & 105217 & 57 \\
    200\_10\_84\_9 & 67.9  & 0.83  & 224551 & 1907  & \textbf{67.4} & 1.02  & 224596 & 1505  & 318   & 0.60  & 117549 & 19    & 318   & 0.00  & \textbf{117543} & 0 \\
    200\_10\_85\_15 & \textbf{62.9} & 0.83  & 303381 & 2050  & 63.2  & 0.60  & 303335 & 2961  & 215   & 0.00  & 195820 & 0     & 215   & 0.00  & 195820 & 0 \\
    200\_20\_145\_15 & 36.6  & 0.80  & 275546 & 3066  & \textbf{36.5} & 0.67  & 277057 & 3948  & 158   & 0.00  & 143507 & 16    & 158   & 0.00  & \textbf{143497} & 0 \\
    200\_20\_150\_9\_D5 & 191.6 & 2.29  & 90882 & 3176  & \textbf{184.8} & 5.02  & 92562 & 1457  & 318   & 16.31 & 51678 & 74    & 345.9 & 1.45  & \textbf{51497} & 2 \\
    200\_20\_54\_15 & \textbf{36.7} & 1.42  & 295455 & 2829  & 37.5  & 0.92  & 293412 & 3656  & 125   & 0.30  & 161424 & 25    & 125   & 0.00  & \textbf{161412} & 0 \\
    200\_20\_55\_9 & \textbf{37} & 1.00  & 229781 & 4000  & 37.7  & 0.78  & 228500 & 5602  & 310   & 8.83  & 71652 & 483   & 328   & 4.96  & \textbf{70154} & 92 \\
    200\_20\_97\_15 & 42    & 0.00  & 287989 & 4572  & \textbf{42}    & 0.00  & 285854 & 5826  & 171   & 0.00  & 156951 & 0     & 171   & 0.00  & 156951 & 0 \\
    200\_20\_97\_9 & \textbf{37.3} & 1.19  & 275710 & 4650  & 37.6  & 0.80  & 276680 & 5627  & 152   & 5.81  & 100450 & 1414  & 168.7 & 0.64  & \textbf{98500} & 43 \\
    200\_40\_130\_9\_D4 & 108   & 0.00  & 103493 & 2383  & \textbf{108}   & 0.00  & 103389 & 1692  & 219   & 4.93  & 48022 & 533   & 216.1 & 0.94  & \textbf{46663} & 329 \\
    200\_40\_133\_15 & 25.4  & 0.66  & 280950 & 4927  & \textbf{25.4}  & 0.66  & 279931 & 3980  & 138   & 4.12  & 98962 & 585   & 145.3 & 3.47  & \textbf{97396} & 72 \\
    200\_40\_45\_15 & \textbf{23.6} & 0.80  & 256232 & 3997  & 24.2  & 0.60  & 256521 & 3155  & 198   & 3.93  & 91369 & 970   & 212.3 & 1.49  & \textbf{87974} & 31 \\
    200\_40\_45\_9 & \textbf{26} & 0.63  & 271406 & 4939  & 26.4  & 0.49  & 267745 & 7041  & 266   & 10.32 & 93099 & 2528  & 301.5 & 9.74  & \textbf{83744} & 1500 \\
    200\_40\_90\_9 & \textbf{25.4} & 0.80  & 292674 & 8765  & 26    & 0.63  & 291293 & 5745  & 219   & 10.62 & 97899 & 3623  & 250.6 & 10.04 & \textbf{85915} & 1533 \\
    200\_40\_91\_15 & \textbf{23.7} & 0.64  & 246065 & 3201  & 24.1  & 0.54  & 248715 & 6059  & 164   & 7.56  & 89262 & 810   & 178.6 & 5.62  & \textbf{86590} & 133 \\ \hline
    \end{tabular}%
  \label{tab:hantco_avg}%
\end{table*}%

All the results presented in tables have been obtained for given ACO parameter configuration: $p=12$, $\mu=0.1$, $p_{init}=1.5$, $\alpha=1$, $\delta=0.05$, $p_{min}=0.05$, $h_{init}=1$, $\beta=0$, $\gamma=150$, $\sigma=30$, $\psi=0.1$, $\kappa_{init}=20$. This configuration has been regarded as the best, defined as a result of the previous parameter--tuning experiments. The same configuration has been chosen to be used in every pheromone update method (ALL, ELITE, DIFF), every optimization mode approach (DO, BO, CO) both for ACO and HAntCO approaches. 

\subsection{Experiments' performance} 

The processing time was varied in relation to the used update method. For ALL method that could be regarded as the simplest, the processing time was relatively small (from 7 to 90 seconds, depending on processed dataset instance). However, for ELITE and DIFF methods, that are regarded as more complex because of the need of sorting ants and choosing best / worst, the processing time varied from 30 to 270 seconds per one execution in one CPU for given parameter configuration\footnote{Machine for tests was equipped with 8 CPUs Intel Core i7 2.67 GHz each, 24 GB of RAM memory and Ubuntu 12.04 OS.}.

\subsubsection{The best found results}

The best results obtained by ACO for CO and DO modes have been compared with the results obtained by using heuristics proposed in \citep{skowronski_heu}. In Tab. \ref{tab:best} this comparison is presented. For each dataset instance and optimization mode, the best results have been chosen from various pheromone update methods. Indication which method provided the best results is stored in columns named $M$ for every optimization mode. 

The obtained best results have been compared with the heuristic results. We decided to omit the name of heuristic if possible to reduce the space covered by the table. For heuristic results in CO, SA heuristic name has been omitted without losing any important information, as the parameter configuration for that method has been written in the table. To give a more detailed view about those methods, please refer to \citep{skowronski_heu}. 

Better values from comparison optimization modes between ACO and heuristics have been written in bold. If key values (duration for DO or cost for CO) were equal for ACO and heuristic approaches, the smallest value of the second aspect has been taken into account to choose a better solution. If both project schedule properties turned out to be the same, both solutions were written in bold. 

To determine the best obtained result for BO mode, neither duration nor cost has been investigated. Instead of those aspects, the evaluation function value has been taken into account. Furthermore, we were not able to compare strictly the results of BO for ACO with corresponding ones for heuristics, as no evaluation function has been used to evaluate results of heuristics.  

A similar analysis has been made for the best found results within investigated hybrid. The best HAntCO results have been presented in Tab. \ref{tab:hantco_best}. The most significant difference for HAntCO best results table in comparison with table of best results for classical ACO is that there is no BO mode included. It is because hybrid is activated only for DO or CO mode -- depending on selected heuristic for initialization. 

Taking into account the results gathered in Tab. \ref{tab:hantco_best}, we can assume that the ELITE strategy mode for HAntCO generally provides better results than DIFF in DO mode. It provided better results in 26 cases (72\%). However, in CO we noticed that the DIFF strategy turned out to be more suitable than the ELITE, provided better results in 9 cases (25\%), while the ELITE became better in only one case (less than 3\%). In remaining cases, both strategies gave the same best results. An interesting fact is that for DO no equal best results for both strategies have been found.    

Also comparing HAntCO best results (see Tab. \ref{tab:hantco_best}) to single heuristics results (see. Tab \ref{tab:best}) we can see that hybrid ACO with heuristics is more effective for DO than CO mode. In most instances (89\%) HAntCO found a better solution than simple heuristic or ACO.

\subsubsection{Averaged results}

Averaged results obtained for various pheromone update methods have been presented in Tab. \ref{tab:average} in a similar way as the ones in Tab. \ref{tab:best} respectively. We also provided in Tab. \ref{tab:average} the notation for the method that provided best results (A, D, E, D$\slash$E). In opposition to Tab. \ref{tab:best} no comparison to averaged heuristic results has been introduced, because heuristics are deterministic methods for which result can be obtained in only one iteration. On the other hand, in Tab. \ref{tab:average} a standard deviation measure ($\sigma$) has been introduced, to indicate the level of variability of the obtained results. It is presented as a percentage value of an average.

For DO and CO modes the smallest averaged values of project duration or project cost respectively have been taken into account to determine the best pheromone update method. If values of given aspect are equal, the smallest value of the second aspect is taken into account. If there is still no possibility to determine which pheromone update method provides better solutions, the standard deviation of more important aspect is taken into account (duration for DO and cost for CO respectively) and the method with smaller standard deviation value is regarded as better.

We have also provided averaged results for HAntCO approach, presented in Tab. \ref{tab:hantco_avg}. Analogously to best \\ HAntCO approach results, averaged ones regard only DO and CO modes. Averaged values are supported by standard deviation values that reflect the variability of the obtained results. We have also decided to count how many times one strategy became better than another also in averaged results. For DO, ELITE strategy became better in 25 cases (69\%), while DIFF became better in the remaining ones. For CO, DIFF strategy provided better results in 14 cases (39\%), while only in one case ELITE strategy became better. For the remaining ones, the obtained averaged results became the same. It leads to conclusion that HAntCO searches space in CO mode in very directed way, being unable to explore other parts of the solution space. Independent character of searching is, in many cases, regardless of applied pheromone update strategy. 

To investigate the level of stability of HAntCO in comparison with classical ACO, we have checked how many times 0--equal standard deviation value has been obtained in the conducted experiments. Those results are presented in Tab. \ref{tab:zero_std}. The results gathered in this table prove that the proposed hybrid approach is more directed and thus, the proposed approach found the same solution in many more cases than classical ACO which stochastic nature allows to explore the search space more widely. 

\begin{table}[!ht]
  \centering
  \caption{Number of 0--equal standard deviation measures for given pheromone update strategies and optimization modes.}
    \tiny \begin{tabular}{|c|r|r|r|r|r|r|r|r|} \hline
    \multirow{3}[0]{*}{Method} & \multicolumn{4}{|c|}{ELITE}     & \multicolumn{4}{|c|}{DIFF}      \\ \cline{2-9}
          & \multicolumn{2}{|c|}{DO} & \multicolumn{2}{|c|}{CO} & \multicolumn{2}{|c|}{DO} & \multicolumn{2}{|c|}{CO} \\ \cline{2-9}
          & \multicolumn{1}{|c|}{Days } & \multicolumn{1}{|c|}{Cost} & \multicolumn{1}{|c|}{Days} & \multicolumn{1}{|c|}{Cost} & \multicolumn{1}{|c|}{Days } & \multicolumn{1}{|c|}{Cost} & \multicolumn{1}{|c|}{Days} & \multicolumn{1}{|c|}{Cost} \\ \hline
    ACO   & 3     & 0     & 5     & 4     & 3     & 0     & 4     & 1 \\
    HAntCO & 2     & 0     & 24    & 21    & 3     & 0     & 25    & 16 \\ \hline
    \end{tabular}%
  \label{tab:zero_std}%
\end{table}%

The most interesting results found in Tab. \ref{tab:zero_std} concern CO mode. For that mode, HAntCO found the same cost solutions 21 (58\%) times for ELITE and 16 times (44\%) for DIFF strategies, while the same duration solutions have been found 24 (67\%) and 25 (69\%) times respectively.

\subsection{Computational complexity}
Our research has been extended by investigating the level of complexity of compared methods. The complexity has been estimated as a number of potential assignments of resources to a given task as dominant operations. As this value is constant regardless of the optimization process and depends only on initial skill constraints, we can compute the level of complexity as a factor of an average number of iterations and a number of possible assignments. The results of those computations are presented in Tab. \ref{tab:dominant}.

As we decided to set a constant number of iterations in most methods like TS, EA S and EA C, the complexity level for those methods was easy to compute. For ACO and HAntCO we decided to get an average number of iterations from all optimization modes (DO, BO, CO) and update pheromone methods (ALL, ELITE, DIFF) as the value that should be multiplied by a number of possible assignments.

\begin{table}[!ht]
  \centering
  \caption{Average number of dominant operations (divided by $10^3$) during optimization process using investigated methods for given parameters' configurations}
    \tiny \begin{tabular}{|r|r|r|r|r|r|r|} \hline
          & D1    & D2    & D3    & D4    & D5    & D6 \\ \hline
    TS    & 200.3  & 38.3  & 22.7  & 234.5 & 159.6 & 72.0 \\
    EA C   & 80.3  & 38,3  & 22.7  & 234.5 & 159.6 & 72.0 \\
    ACO   & 1287.9 & 472.8 & 205.9 & 3038.4 & 2221.4 & 1063.4 \\
    HAntCO & 423.9 & 212.5 & 86.2  & 1925.5 & 1481.2 & 323.3 \\
    H     & 0.803 & 0.383 & 0.227 & 2.345 & 1.596 & 0.72 \\ \hline
    \end{tabular}%
  \label{tab:dominant}%
\end{table}%

Based on the results gathered in Tab.\ref{tab:dominant} we can notice that ACO and HAntCO are most processing--complex methods. However, the level of complexity for HAntCO is lower than for classical ACO. It is because the number of iterations for hybrids is generally smaller, as searching is started from more directed place in the solution space. 

Complexity level of heuristics has been computed as multiplication of a number of possible assignments by 1, as there is only one iteration in heuristic scheduling process. What is more, heuristic are deterministic approaches. Therefore, we always get the same results that are obtained in only one iteration. Hence, heuristic could be used as a powerful tool to get the first glance of optimization possibilities for given dataset instance. 

\subsection{Results' discussion}

Both for the best and averaged results for classical ACO, ELITE update pheromone method turned out to be the best for DO mode, while DIFF update pheromone method became the most suitable for BO mode. However it is not possible to get such straightforward conclusions for CO mode, because DIFF method became the most suitable choice for the best obtained results while both DIFF and ELITE methods provided equally good results for average obtained optimization results. 

We have also compared pheromone update methods in hybrids performance. For that approach ELITE mode turned out to be the most suitable for DO, while any ($*$) proposed pheromone update method became equally good for CO mode for most project instances. No difference between pheromone update method has been also observed in 15/36 (42\%) cases in CO. It could lead to conclusion that pheromone update method is not as crucial as for classical ACO. It is because initial solution is preferred -- hybrid is more exploitation-- than exploration--oriented. 

We have also compared how many times heuristics provided better results than the best ones obtained from an application of ACO approaches (see Tab. \ref{tab:best}). For DO, SLS heuristic became better 9 times (25\%), while for CO SA or RS heuristics became better 18 times (50\%). It shows that classical ACO approach, proposed in this paper cannot be fully regarded as better in comparison with heuristic methods. However, combining it with heuristic in hybrid (HAntCO) approach turned out to give usually much better results than any other investigated methods, especially for CO mode. 

An interesting fact is that DO mode is generally more stable than other based on the provided results. It has been deducted by counting number of bigger than 10\% $\sigma$ values in Tab. \ref{tab:average}. For DO there were no such values, while, for BO, there were 3 over 10\% values (2 for duration aspect and 1 for a cost aspect). Finally, for CO, there were 7 over 10\% values of standard deviation -- all for a duration aspect.

An interesting conclusion that could be made regardless of the best or averaged results is that a DIFF strategy provided better solutions in DO mode but mostly for dataset instances containing 200 tasks. The best results obtained by a DIFF strategy were better than obtained by an ELITE in 9 cases for 200 task--project instances (50\%), while ELITE strategy provided only one better solution than a DIFF (5\%). Averaged results obtained in a DIFF mode were better in 12 cases (67\%), while ELITE strategy still provided only one better solution in comparison with a DIFF.  

Comparing the best results obtained by ACO and HAnt-CO it can be noticed that HAntCO outclasses classical ACO, whichever pheromone update method would be used. For DO, classical ACO approach has been better than HAntCO in only 5 from 36 cases, while for CO HAntCO became better than ACO for every project instance. Analysing averaged results, there are only 3 cases with ACO results better than HAntCO ones. Still only for DO. For CO, ACO has never been better than HAntCO. It proves the legitimacy of using hybrids that become robust way of boosting optimization process. 

To get bigger awareness of classical ACO and HAnt-CO approaches' robustness, we decided to compare the obtained best results for ACO with best results obtained using other methods, as EA  \citep{skowronski_men} and TS \citep{skowronski_ts}. However, we needed to distinguish the best results obtained for DO and CO modes from BO mode, because no heuristic scheduling method has been proposed for BO. Comparison of DO, BO and CO modes has been presented in Tab. \ref{tab:comparison}.

\begin{table*}[!ht]
  \centering
  \caption{Comparison of best obtained results for investigated methods in DO, BO and CO modes}
    \begin{tabular}{|c|c|r|r|r|r|r|r|r|} \hline
    Method & Mode  & \multicolumn{1}{|c|}{Crit.} & \multicolumn{1}{|c|}{D1} & \multicolumn{1}{|c|}{D2} & \multicolumn{1}{|c|}{D3} & \multicolumn{1}{|c|}{D4} & \multicolumn{1}{|c|}{D5} & \multicolumn{1}{|c|}{D6} \\ \hline
    \multirow{6}[0]{*}{TS} & \multirow{2}[0]{*}{DO} & days  & 32    & 33    & 51 & \textbf{92} & 179   & 199 \\
          &       & cost  & 40656 & 43542 & 40054 & 88720 & 80448 & 97978 \\ \cline{2-9}
          & \multirow{2}[0]{*}{BO} & days  & 37    & 49    & 61    & 125   & 184   & 222 \\
          &       & cost  & 38939 & 34240 & 36100 & 50438 & 54181 & 75996 \\ \cline{2-9}
          & \multirow{2}[0]{*}{CO} & days  & 129   & 179   & 133   & 254   & 481   & 330 \\
          &       & cost  & 30750 & 26444 & 31645 & 46371 & 52425 & 73126 \\ \hline
    \multirow{6}[0]{*}{EA S } & \multirow{2}[0]{*}{DO} & days  & 32    & 34    & 52    & 112   & 179   & 216 \\
          &       & cost  & 41509 & 42804 & 40768 & 66196 & 90753 & 81344 \\ \cline{2-9}
          & \multirow{2}[0]{*}{BO} & days  & 32    & 40    & 57    & 112   & 188   & 216 \\
          &       & cost  & 42975 & 40387 & 38486 & 87107 & 84067 & 88317 \\ \cline{2-9}
          & \multirow{2}[0]{*}{CO} & days  & 116   & 133   & 163   & 196   & 417   & 294 \\
          &       & cost  & 30158 & 26691 & 34361 & 52027 & 52400 & 74897 \\ \hline
    \multirow{6}[0]{*}{EA C } & \multirow{2}[0]{*}{DO} & days  & 35    & 52    & 64    & 112   & 183   & 216 \\
          &       & cost  & 41217 & 37248 & 40242 & 87487 & 81555 & 99462 \\ \cline{2-9}
          & \multirow{2}[0]{*}{BO} & days  & 46    & 77    & 77    & 114   & 211   & 216 \\
          &       & cost  & 37190 & 31888 & 35527 & 79854 & 72918 & 92602 \\ \cline{2-9}
          & \multirow{2}[0]{*}{CO} & days  & 56    & 94    & 84    & 120   & 230   & 216 \\
          &       & cost  & 35760 & 31328 & 34160 & 78928 & 72338 & 91972 \\ \hline \hline
    \multirow{6}[0]{*}{ACO} & \multirow{2}[0]{*}{DO} & days  & 32    & 34    & \textbf{50}    & 112   & 186   & 216 \\
          &       & cost  & 52119 & 44999 & 41029 & 94488 & 91461 & 99375 \\ \cline{2-9}
          & \multirow{2}[0]{*}{BO} & days  & 60    & 72    & 75    & 132   & 247   & 237 \\
          &       & cost  & 30900 & 27120 & 31681 & 48362 & 52620 & 72753 \\ \cline{2-9}
          & \multirow{2}[0]{*}{CO} & days  & 121   & 129   & 167   & 205   & 296   & 274 \\
          &       & cost  & 30107 & \textbf{26323} & \textbf{30164} & 48419 & 51678 & 72036 \\ \hline
    \multirow{4}[0]{*}{HAntCO} & \multirow{2}[0]{*}{DO} & days  & \textbf{23} & \textbf{33} & 53    & 108   & \textbf{177} & \textbf{186} \\
          &       & cost  & 53358 & 42199 & 40811 & 104965 & 92567 & 103561 \\ \cline{2-9}
          & \multirow{2}[0]{*}{CO} & days  & 117   & 129   & 167   & 216   & 344   & 267 \\
          &       & cost  & \textbf{30104} & \textbf{26323} & \textbf{30164} & 46342 & 51496 & 71986 \\ \hline \hline
    \multirow{4}[0]{*}{H } & \multirow{2}[0]{*}{DO} & days  & 32    & 38    & 57    & 112   & 183   & 216 \\
          &       & cost  & 53154 & 44309 & 40539 & 101879 & 92821 & 105593 \\ \cline{2-9}
          & \multirow{2}[0]{*}{CO} & days  & 119   & 129   & 167   & 214   & 337   & 256 \\
          &       & cost  & 30104 & \textbf{26323} & \textbf{30164} & \textbf{46133} & \textbf{51496} & \textbf{71986} \\ \hline
    \end{tabular}%
  \label{tab:comparison}%
\end{table*}%

This comparison has been made only for project instances D1--D6, because only those have been investigated in \citep{skowronski_men}, \citep{skowronski_ts} and \citep{skowronski_heu}. The compared methods are Taboo Search (TS), specialized Evolutionary Algorithms (EA S), classic EA (EA C), classical ACO, HAntCO and heuristics (H). 


\begin{table*}[!ht]
  \centering
  \caption{Comparison of averaged obtained results for investigated methods in DO and CO modes}
    \begin{tabular}{|r|r|r|r|r|r|r|r|r|} \hline
    Method & Mode  & Crit. & D1    & D2    & D3    & D4    & D5    & D6 \\ \hline
    \multicolumn{1}{|c|}{\multirow{4}[0]{*}{TS }} 
& \multicolumn{1}{|c|}{\multirow{2}[0]{*}{DO}} & days  & 35.06$\pm$2.26 &	46.14$\pm$3.06
&	71.0$\pm$0.0 &	112$\pm$0.0 &183.0$\pm$0.0 &	216.0$\pm$0.0\\ 
    \multicolumn{1}{|c|}{} & \multicolumn{1}{|c|}{} 
    & cost  & 41151$\pm$201 &	38205$\pm$950 &	38748$\pm$0.0 &	87691$\pm$206 & 79927$\pm$166 & 98538$\pm$138 \\ \cline{2-9}

    \multicolumn{1}{|c|}{} & \multicolumn{1}{|c|}{\multirow{2}[0]{*}{CO}} & days  
& 128$\pm$4.99 &	176.7$\pm$11.6 &	133.4$\pm$4.4	& 248.3$\pm$21.4 &	467.3$\pm$23.7 &	358.2$\pm$17.2 \\ 
    \multicolumn{1}{|c|}{} & \multicolumn{1}{|c|}{} & cost  
& 30693$\pm$2.1 &	26424$\pm$3.4 &	31637$\pm$0.0 &	46359$\pm$128 &	52354$\pm$43 & 72961$\pm$0.0
 \\ \hline
\multicolumn{1}{|c|}{\multirow{4}[0]{*}{EA S }} & \multicolumn{1}{|c|}{\multirow{2}[0]{*}{DO}} & days  

& 32$\pm$0.00 &	37.52$\pm$1.28 &	54.68$\pm$1.39	& 112$\pm$0.00 &	\textbf{180$\pm$1.51} &	216$\pm$0.00
 \\ 
    \multicolumn{1}{|c|}{} & \multicolumn{1}{|c|}{} & cost  
& 52781$\pm$1510  & 43547$\pm$909  & 41082$\pm$544  & 104459$\pm$4194  &	92355$\pm$3234  &	100002$\pm$4511  \\ \cline{2-9}
    \multicolumn{1}{|c|}{} & \multicolumn{1}{|c|}{\multirow{2}[0]{*}{CO}} & days  
& 43.9$\pm$7.64 & 150 $\pm$3.09 & 	110.7$\pm$10 & 	234.66$\pm$20.4 & 	443.8$\pm$25.8 & 	221.6.5$\pm$10.88 \\ 
    \multicolumn{1}{|c|}{} & \multicolumn{1}{|c|}{} & cost  
& 46492$\pm$673& 26344$\pm$57 & 	34834$\pm$535 & 	47600$\pm$509 & 	\textbf{51200$\pm$220} & 	93914$\pm$957  \\ \hline

\multicolumn{1}{|c|}{\multirow{4}[0]{*}{EA C }} & \multicolumn{1}{|c|}{\multirow{2}[0]{*}{DO}} & days  
& 32.0$\pm$0.00 &	46.6$\pm$2.27 &	68.32$\pm$1.72 &	111.88$\pm$0.72 &	181.2$\pm$1.48 &	216.0$\pm$0.00  \\ 

    \multicolumn{1}{|c|}{} & \multicolumn{1}{|c|}{} & cost  
& 52949$\pm$1850 &	43113$\pm$1139 &	41026$\pm$927 &	107021$\pm$2955 &	87899$\pm$2687 &	101798$\pm$1894  \\ \cline{2-9}

    \multicolumn{1}{|c|}{} & \multicolumn{1}{|c|}{\multirow{2}[0]{*}{CO}} & days  
& 46.43$\pm$5.84 &	76.45$\pm$7.29&	114.2$\pm$12.11 &	116.5$\pm$5.9 &	206.36$\pm$12.07 &	219.34.7$\pm$6.97 \\ 
    \multicolumn{1}{|c|}{} & \multicolumn{1}{|c|}{} & cost  
& 45220$\pm$902 &	36678$\pm$656 &	34074$\pm$521 &	94577$\pm$1586 &	77804$\pm$1228 &	94218$\pm$852 \\ \hline

\hline
\multicolumn{1}{|c|}{\multirow{4}[0]{*}{ACO}} & \multicolumn{1}{|c|}{\multirow{2}[0]{*}{DO}} & days  &

 32$\pm$0.0 &	38.4$\pm$1.49 &	\textbf{52.86$\pm$1.6} &	112$\pm$0.0 &	189.8$\pm$2.5 &	216.24$\pm$0.72  \\ 
    \multicolumn{1}{|c|}{} & \multicolumn{1}{|c|}{} & cost  & 
53092$\pm$1816 &	43271$\pm$895 &	53092$\pm$1816 &	104862$\pm$2928 &	90471$\pm$2765 &	102075$\pm$1930 \\ \cline{2-9}

    \multicolumn{1}{|c|}{} & \multicolumn{1}{|c|}{\multirow{2}[0]{*}{CO}} & days  & 
114.06$\pm$7.29 &	127.5$\pm$6.5 &	166.82$\pm$0.38 &	181.52$\pm$12.62 &	252.9$\pm$11.93 &	260.35$\pm$8.27  \\ 
    \multicolumn{1}{|c|}{} & \multicolumn{1}{|c|}{} & cost  & 
30295$\pm$332 &	26376$\pm$154 &	30167$\pm$7.21 &	50486$\pm$1113 &	53110$\pm$584 &	72767$\pm$1566 \\ \hline

\multicolumn{1}{|c|}{\multirow{4}[0]{*}{HAntCO}} & \multicolumn{1}{|c|}{\multirow{2}[0]{*}{DO}} & days  &

 \textbf{25.1$\pm$0.81} &	\textbf{35.8$\pm$1.07} &	55.8$\pm$0.73 &	\textbf{108.0$\pm$0.00} &	182.48$\pm$5.05 &	\textbf{186.8$\pm$2.16} \\ 
    \multicolumn{1}{|c|}{} & \multicolumn{1}{|c|}{} & cost  
& 53527$\pm$1086 &	44183$\pm$622 &	56671$\pm$314 &	104112$\pm$2217 &	90294$\pm$3198 &	104510$\pm$1690  \\ \cline{2-9}

    \multicolumn{1}{|c|}{} & \multicolumn{1}{|c|}{\multirow{2}[0]{*}{CO}} & days  
& 117.0$\pm$0.00 &	128.98$\pm$0.13 &	167.0$\pm$0.00 &	217.1$\pm$1.07 &	341.62$\pm$8.02 &	267.36$\pm$1.94 \\ 
    \multicolumn{1}{|c|}{} & \multicolumn{1}{|c|}{} & cost  
& \textbf{30104$\pm$0.0} &	\textbf{26323$\pm$3.78} &	\textbf{30164$\pm$0} &	46554$\pm$291 &	51514$\pm$ &	\textbf{71986$\pm$0.00} \\ \hline
   \hline

    \multicolumn{1}{|c|}{\multirow{4}[0]{*}{H }} & \multicolumn{1}{|c|}{\multirow{2}[0]{*}{DO}} & days  & 32$\pm$0    & 38$\pm$0    & 57$\pm$0    & 112$\pm$0   & 183$\pm$0   & 216$\pm$0 \\ 
    \multicolumn{1}{|c|}{} & \multicolumn{1}{|c|}{} & cost  & 53154$\pm$0 & 44309$\pm$0 & 40539$\pm$0 & 101879$\pm$0 & 92821$\pm$0 & 105593$\pm$40 \\ \cline{2-9}
    \multicolumn{1}{|c|}{} & \multicolumn{1}{|c|}{\multirow{2}[0]{*}{CO}} & days  & 119$\pm$0   & 129$\pm$0   & 167$\pm$0   & 214$\pm$0   & 337$\pm$0   & 256$\pm$0 \\ 
    \multicolumn{1}{|c|}{} & \multicolumn{1}{|c|}{} & cost  & \textbf{30104$\pm$0} & \textbf{26323$\pm$0} & \textbf{30164$\pm$0} & \textbf{46133$\pm$0} & 51496$\pm$0 & \textbf{71986$\pm$0} \\ \hline
    \end{tabular}%
  \label{tab:CDComparison}%
\end{table*}%

\begin{table*}[!ht]
  \centering
  \caption{Results of the unpaired t-test between the best and the second best performing methods (for each instances D1-D6) based on Tab.9 (heuristic H \citep{skowronski_heu} results not included as a part of HAntCO)}

   \begin{tabular}{|r|r|r|r|r|r|r|r|} \hline
			      instance  & mode & best methods & std.error & t    & 95\% conf. inter.   & two tailed p    & stat. significance     \\ \hline

 \hline
    \multirow{2}{*}{D1}    		& DO & \textbf{HAntCO}, EA S 
&  0.115 & 60.2350  & -7.12 to -6.67  & $<0.0001$ & \textbf{extr. significant} \\
					& CO & \textbf{HAntCO}, EA S 
& 47.016  & 4.0574 & -284.06 to -97.45 &  $<0.0001$ & \textbf{extr. significant}\\ \hline

     \multirow{2}{*}{D2}    		&  DO & \textbf{HAntCO}, EA S  
&  8.072 & 7.2901 & -2.18 to -1.25 & $<0.0001$ & \textbf{extr. significant}\\ 
					& CO  & \textbf{HAntCO}, TS 
&  0.726  & 2.6885 &   -37.71 to -5.68 &  0.0084 & \textbf{very significant}\\  \hline

    \multirow{2}{*}{D3}    		&  DO & \textbf{ACO}, EA S      
&  0.300  &  6.0720 &  -2.41 to -1.22  & $<0.0001$ & \textbf{extr. significant}\\
					& CO  & \textbf{HAntCO}, \textbf{ACO}  
& 1.018  & 3.5355 & 1.57 to 5.62 &  0.0006 & \textbf{extr. significant}\\ \hline

    \multirow{2}{*}{D4}   		&  DO& \textbf{HAntCO}, EA C
 & 0.102   &  38.1052 &  3.67 to 4.08 & $<0.0001$ & \textbf{extr. significant}  \\
					& CO  & TS, \textbf{HAntCO} 
& 44.934 & 4.3397 & -284.16 to -105.83  & $<0.0001$ & \textbf{extr. significant}\\ \hline

      \multirow{2}{*}{D5}    		&  DO & EA S, EA C
 & 0.299  & 2.8761 & -1.45 to -0.26 &  0.0049 &  \textbf{very significant} \\
					& CO  & EA S, \textbf{HAntCO}  
&  31.673  & 9.8913 & -376.14 to -250.43 &  $<0.0001$ & \textbf{extr. significant} \\ \hline

    \multirow{2}{*}{D6}    		&  DO & \textbf{HAntCO}, TS, EA 
& 0.305   & 95.8523  & 28.67 to 29.88  & $<0.0001$ & \textbf{extr. significant} \\
					& CO  & \textbf{HAntCO}, \textbf{ACO} 
& 221.542 &  3.5243 & 341.14 to 1220.43  & 0.0006 & \textbf{extr. significant} \\ \hline

   \end{tabular}%
  \label{tab:STATScomparision}%
\end{table*}%


The results presented in Tab. \ref{tab:comparison} show that both HAnt-CO and TS outclassed other methods in DO mode, obtaining best cost results for half of investigated project instances for each method (D1, D2, D5, D6 for HAntCO and D3, D4 for TS). For CO mode, classical ACO became the best approach for D2 and D3 instances, while HAntCO obtained the best results for the same instances plus D1. However, the most successful approach for these instances is a heuristic one that allowed to get best results in 5/6 cases. 

The averaged results of investigated methods are presented in Tab.\ref{tab:CDComparison}. It differs slightly from the results in Tab.\ref{tab:comparison}, as methods are non--deterministic. However, conclusions are very similar: HAntCO outperforms other methods in almost every case or results are comparable. We developed extra statistical analysis to prove a quality of presented method. We have provided the Kolmogorov--Smirnov (K--S test) to investigate the normality of the distribution of gained results. The K--S test proved that results of used methods are normally distributed and t--test can be used. Moreover, a sample size around 50 allows the normality assumptions conducive for performing the unpaired t-tests \citep{flury}. We used two tailed t--test with 95\% confidence interval (see results in Tab.\ref{tab:STATScomparision}) for the best and the second best performing methods applied in D1--D6 instances for DO and CO modes.
 
We found that HAntCO results are the best in most cases.
Very interesting results are noticed for EA S, especially for D5 instance (in DO and CO mode) where EA S gives the best (average) solutions.
Only in one case (D3 instance DO mode) ACO gives better average solution. The results are significant in statistical meaning. The statistical significance of results for HAntCO in CO mode comes mostly from the fact that HAntCO is a method directed by a heuristic that finds the best cost--oriented solution (algorithm). Hence, the statistical significance of this method should be mostly investigated in DO mode. In this mode, the results obtained by HAntCO are statistically significant in 3 cases (D1, D2, D6), while DO--oriented results obtained by ACO are statistically significant in only one case (D3). It additionally proves the legitimacy of using proposed hybrid rather than classical ACO approach.
We have also investigated results for several methods in BO mode. In this case, classical ACO approach outclassed the rest of examined methods and became the best choice in 5/6 cases. However, it caused enlarging the project schedule duration of analysed instances and make them mostly the longest from all obtained with various methods. EA with specialized genetic operators gave the smallest project cost for BO mode. It was the best in 5/6 cases. An interesting fact is that the results obtained for ACO are completely different from the results from other methods like TS or EA. The conclusion could be that ACO searches the solution space totally different from the above--mentioned methods. Hence combining those approaches into one could be possibly effective and potentially give promising results. 


\section{Conclusions and further work \label{sec:conclusions}}
 
In this paper we have presented hybrid approach for solving Multi--Skill Resource--Constrained Project Sche-duling Problem. MS--RCPSP is an extension of classical RCPSP with skills and cost domain. Our approach bases on classical ACO metaheuristics for discrete combinatorial problems. However, it has been enhanced by modified pheromone update methods. Furthermore, we have proposed a hybridization of ACO approach (HAnt-CO) by using simple heuristics based on priority rules to find an initial solution in optimization process. 

What is more, we have prepared and published iMOP-SE dataset instances to allow others to investigate their approaches for such defined MS--RCPSP. The dataset consists of 36 instances containing 100 or 200 tasks. All instances are varied by the number of resources, precedence relations and skill types what makes them more or less difficult to be scheduled. 

We have also defined evaluation methods for the proposed approaches not only in case of their robustness (how good the final solution is) but also their effectiveness. To evaluate method's quality, we rate not only the project schedule duration, as in classical RCPSP, but also the project schedule performance cost, regarding the MS--RCPSP as multi--objective optimization problem. The method's effectiveness is rated by the number of dominant operations that need to be performed during the optimization process.

Finally, we have compared the results obtained by HAnt-CO and ACO with the ones received with the use of other methods as simple heuristics, Taboo Search and Evolutionary Algorithms with classic and specialized genetic operators that have been published before. The provided results have been also supported by the statistical significance tests. The obtained results lead to the conclusion that ACO--based approaches stand suitable ones for solving MS--RCPSP as they provide mostly the best results from all investigated methods. 

\subsection{Future work}

After observation that pheromone update method in ACO has an impact on the obtained results depending on selected optimization mode (aspect), we are encouraged to use this outcome and propose an approach more dedicated to multi--objective optimization using Pareto front from various ant populations performing in different pheromone update methods. It could provide us a mechanism to find very good solutions leaving the need of setting optimization mode. It could give us good solutions in DO and BO in the same run of ACO--based run. 

Pareto--based approach could be implemented in the investigated methods to distinguish non--dominated solutions. By non--dominated solution a one with the smallest value of given criterion is taken while remaining criterion values are equal. In MS--RCPSP non--dominated solution is regarded as the one that has smallest cost or duration from subset of solutions with the same duration or cost respectively. It could make the optimization process more robust and effective, as \textit{good enough} results could be found in a smaller number of iterations within the examined method. 

As cost oriented optimization in ACO and HAntCO has not provided significantly better results than other methods investigated in this paper, we discuss a potential application of dedicated neighbourhood definition for ants to make them more oriented to search solutions cheaper. 

Investigating the comparison of the results obtained for CO and DO modes could lead to conclusion that ACO is a powerful tool for solving MS--RCPSP, especially if it was boosted by initial solution obtained by heuristic (HAntCO). It leads to conclusion that other hybrids should be investigated using the proposed heuristics. Hence we would examine and compare the results obtained for EA, TS or SA approaches to check, whether boosting initial solution by heuristic provides better results for other metaheuristics. 

According to the experiences with ACO of other researchers, ACO can be extended by additional heuristic \citep{dorigo} to enhance the potential of optimization. We plan to find suitable, problem specific heuristic that could be used and investigate whether it could make our approach better in solving MS--RCPSP.

\end{document}